\pdfoutput=1

\documentclass[11pt]{article}

\usepackage{style/emnlp2023}

\usepackage{times}
\usepackage{latexsym}
\usepackage{siunitx}
\usepackage[T1]{fontenc}

\usepackage[utf8]{inputenc}

\usepackage{microtype}

\makeatletter
\renewcommand{\paragraph}{%
  \@startsection{paragraph}{4}%
  {\z@}{1ex \@plus 1ex \@minus .2ex}{-1em}%
  {\normalfont\normalsize\bfseries}%
}
\makeatother

\newcommand{\representative}{\textit{representative}}
\definecolor{Gray}{gray}{0.85}

\usepackage{makecell} 
\usepackage{algorithm}
\usepackage[noend]{algpseudocode}
\usepackage{varwidth}
\usepackage{multirow}
\usepackage{tabularx}
\usepackage{subcaption}
\usepackage{amssymb}
\usepackage{amsmath}
\usepackage{amsthm}
\usepackage{amsfonts}
\usepackage{bbm}
\usepackage{bm}
\usepackage{booktabs}
\usepackage{graphicx}
\usepackage{multirow}
\usepackage{multicol}
\usepackage{enumitem}
\usepackage{mdwlist}
\usepackage{csquotes}
\usepackage{multirow}
\include{cv-commands}
\setlist{nosep}

\title{Not all Fake News is Written: \\ A Dataset and Analysis of Misleading Video Headlines}

\author{Yoo Yeon Sung \\
    University of Maryland \\ 
    \texttt{yysung53@umd.edu} \And
    Jordan Boyd-Graber \\
    University of Maryland \\
    \texttt{jbg@umiacs.umd.edu} \\\And
    Naeemul Hassan\\
    University of Maryland\\
    \texttt{nhassan@umd.edu}}

\date{}

\newif\ifcomment\commenttrue

\usepackage[a-1b]{pdfx}

\usepackage{framed}
\usepackage{mdwlist}
\usepackage{siunitx}
\usepackage{latexsym}
\usepackage{colortbl}
\usepackage{xcolor}
\usepackage{nicefrac}
\usepackage{booktabs}
\usepackage{fnpct}
\usepackage{amsfonts}
\usepackage[T1]{fontenc}
\usepackage{bold-extra}
\usepackage{amsmath}
\usepackage{amssymb}
\usepackage{bm}
\usepackage{graphicx}
\usepackage{mathtools}
\usepackage{microtype}
\usepackage{multirow}
\usepackage{multicol}
\usepackage{xpatch}
\usepackage{latexsym,comment}
\usepackage[normalem]{ulem}

\newcommand*{\missingreference}{{\Huge \colorbox{red}{?reference?}}}
\newcommand*{\missingcitation}{{\Huge \colorbox{red}{?citation?}}}

\makeatletter
\xpatchcmd{\@setref}{\bfseries}{\missingreference}{}{}
\def\@citex[#1]#2{\leavevmode
    \let\@citea\@empty
    \@cite{\@for\@citeb:=#2\do
        {\@citea\def\@citea{,\penalty\@m\ }%
            \edef\@citeb{\expandafter\@firstofone\@citeb\@empty}%
            \if@filesw\immediate\write\@auxout{\string\citation{\@citeb}}\fi
            \@ifundefined{b@\@citeb}{\hbox{\reset@font\missingcitation}%
                \G@refundefinedtrue
                \@latex@warning
                {Citation `\@citeb' on page \thepage \space undefined}}%
            {\@cite@ofmt{\csname b@\@citeb\endcsname}}}}{#1}}
\makeatother

\newcommand{\gem}[1]{\mbox{\textsc{gem}}}
\newcommand{\abr}[1]{\textsc{#1}}

\newcommand{\hidetext}[1]{}
\newcommand{\ignore}[1]{}

\ifcomment
    \newcommand{\pinaforecomment}[3]{\colorbox{#1}{\parbox{.8\linewidth}{#2: #3}}}

    \newcommand{\prtodo}[1]{\pinaforecomment{lightblue}{pr}{#1}}
    \newcommand{\prtodoi}[1]{\pinaforecomment{lightblue}{pr}{#1}}
\else
    \newcommand{\pinaforecomment}[3]{}
    \newcommand{\prtodo}[1]{}
    \newcommand{\prtodoi}[1]{}
\fi

\newcommand{\smallurl}[1]{ \begin{tiny}\url{#1}\end{tiny}}

\definecolor{lightblue}{HTML}{3cc7ea}
\definecolor{CUgold}{HTML}{CFB87C}
\definecolor{grey}{rgb}{0.95,0.95,0.95}
\definecolor{ceil}{rgb}{0.57, 0.63, 0.81}
\definecolor{UMDred}{HTML}{ed1c24}
\definecolor{UMDyellow}{HTML}{ffc20e}

\usepackage{pifont}
\usepackage{arydshln}
\usepackage[normalem]{ulem}

\newcommand{\name}{\abr{vmh}}

\graphicspath{ {figures/}{auto_fig/} }

\begin{document}
\maketitle
\begin{abstract}
Polarization and the marketplace for impressions have conspired to
make navigating information online difficult for users, and while
there has been a significant effort to detect false or misleading text,
multimodal datasets have received considerably less attention.
To complement existing resources, we present multimodal Video
Misleading Headline (VMH), a dataset that consists of videos and
whether annotators believe the headline is representative of the
video's contents.
After collecting and annotating this dataset, we analyze multimodal
baselines for detecting misleading headlines.
Our annotation process also focuses on \emph{why} annotators view a
video as misleading, allowing us to better understand the interplay of
annotators' background and the content of the videos.

\end{abstract}
\section{Introduction}
\begin{table}[t]
    \renewcommand{\arraystretch}{1.0}
    \small
    \centering
    \scalebox{0.9}{
    \begin{tabularx}{\linewidth}{lX}
    \Xhline{1.5pt}  
    \rowcolor{Gray} \multicolumn{2}{c}{\textbf{VMH Dataset}}   \\ 
    \Xhline{1.5pt}  \\
        \textbf{Headline}          & \textbf{Clinton Says Trump ``Making Up Lies'' About \colorbox{red!30}{New FBI Review}} \\[0.8ex]
         \textbf{Video}            & \url{https://www.facebook.com/watch/?v=10154955844338812} \\[0.8ex] 
        \textbf{Label}            & \colorbox{red!30}{\textbf{Misleading}}\\
         \textbf{Rationale}          & \textbf{The headline \colorbox{red!30}{implies more than} what is introduced in the video.} \\[0.8ex]
         \textbf{Subrationale}      & \textbf{The headline \colorbox{red!30}{exaggerates} the video content.} \\[1.8ex]
         \cdashline{1-2} \\
         \textbf{Annotator ID}          & \texttt{A2P8V5SKYLL5I4} \\[0.8ex]
         \textbf{Annotator Profile} & \texttt{Ages 30-49, Black, Democratic, Men, Post college} \\[0.8ex]
         \textbf{Venue}          & ABC News\\[0.8ex]
         \textbf{Venue Kind}          & Broadcast   \\[0.8ex]
         \textbf{Venue Credibility}          & High\\[0.8ex]
         \textbf{News Topic}          &  Politics\\[0.8ex]
         \textbf{Headline Property}          & Factual Statement\\[0.8ex]
         \textbf{Transcript}         & \dots is already making up lies about this he is doing his best to confuse  misleading and discourage the American people \\[0.8ex]\Xhline{1.5pt}  

    \end{tabularx}}  
    \caption{\name{} includes video headline, video, annotator's label, and rationales the label is grounded. In the video, the part about ``New FBI Review'' was not present, and thereby annotation is \emph{misleading} because the headline was implying more than the video content.}
    \vspace{-0.5cm}
    \label{table:dataset_example}
\end{table}

\begin{figure*}[!t]
    \centering
    \includegraphics[width=\linewidth]{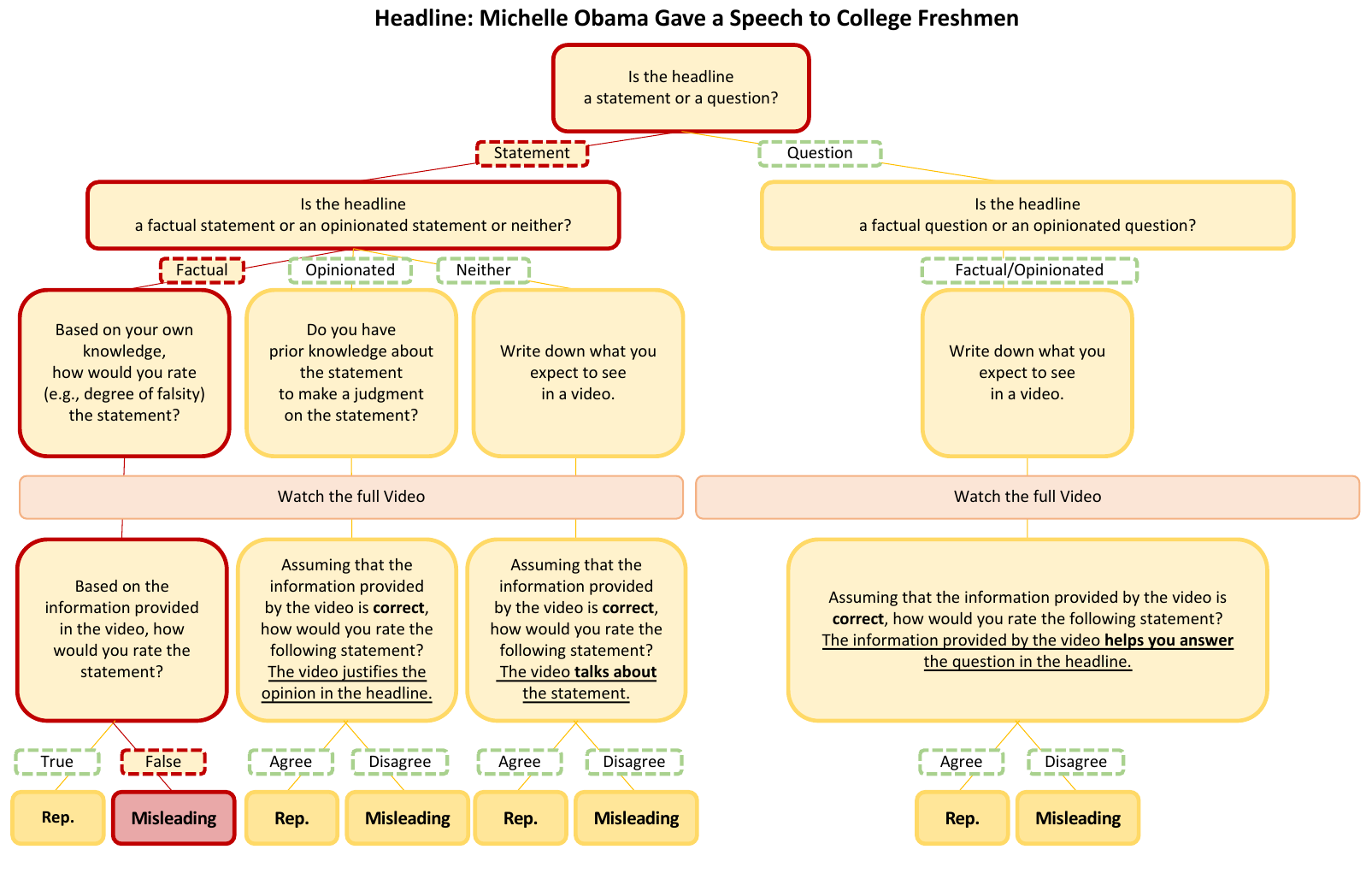} 
    \caption{In the annotation tree, the annotators first consider if the headline ``Michelle Obama Gave a Speech to College Freshmen" is a factual statement. Next, they answer the question, ``Based on the information provided in the video, how would you rate the statement?'' Because the answer was \textit{False}, the implied label is
    \textit{misleading}. The headline is indeed \emph{misleading} because whether ``College Freshmean'' were present in the video is unclear, making it impossible to assess the veracity. Rep.\ refers to \textit{representative} label.}
    \label{fig:diagram}
\end{figure*}

Social media platforms are used by half of \abr{us} adults for everyday
news consumption~\citep{walker}. They have supplanted television as the most common purveyor of
news \citep{wakefield}.
However, content created on these online platforms are often lower
quality than traditional sources and more prone to false stories.
\citet{vosoughi2018spread} contend that false news spreads six times
faster online than offline.

This work focuses on one part of this problem: does a video
headline match its content.
We call this  {\bf misleading video headline} detection.
In text, this is called incongruent headline
detection~\citep{chesney2017incongruent} and is an important problem
because the headline is the first step to a reader accessing
content~\citep{dos2015breaking}.
While there has been work to automatically detect misleading headlines from text
(Section~\ref{related}), users are more likely to believe
fake news when it is accompanied by videos
\citep{wang2021seeing}---and there are no datasets to train models for
misleading video headline detection. 

Hence, it is necessary to investigate content
outside the text (e.g., with videos) as it can help make a more
informed decision by directly analyzing the relationship between the
headline and the video.

To understand this new task, we create a new dataset\footnote{https://github.com/yysung/VMH/tree/master}---Video
Misleading Headline (\name{})---that includes $2{,}247$ news articles
labeled as \textit{representative} or \textit{misleading} (Section~\ref{VMH}).
A careful annotation process captures not just whether videos are misleading but \emph{why}, with specific rationales. We further investigate videos, label rationales, and headline meta information (e.g., venues, news topics, and headline properties) to analyze the features that may contribute towards
an instance being identified as misleading
(Section~\ref{section3}).
Section~\ref{Experiments} shows that existing models fail to identify
misleading video headlines, showing that this important but difficult
task requires further research in both the text and visual domains.

\begin{figure}[!t]
    \centering
    \includegraphics[width=\linewidth]{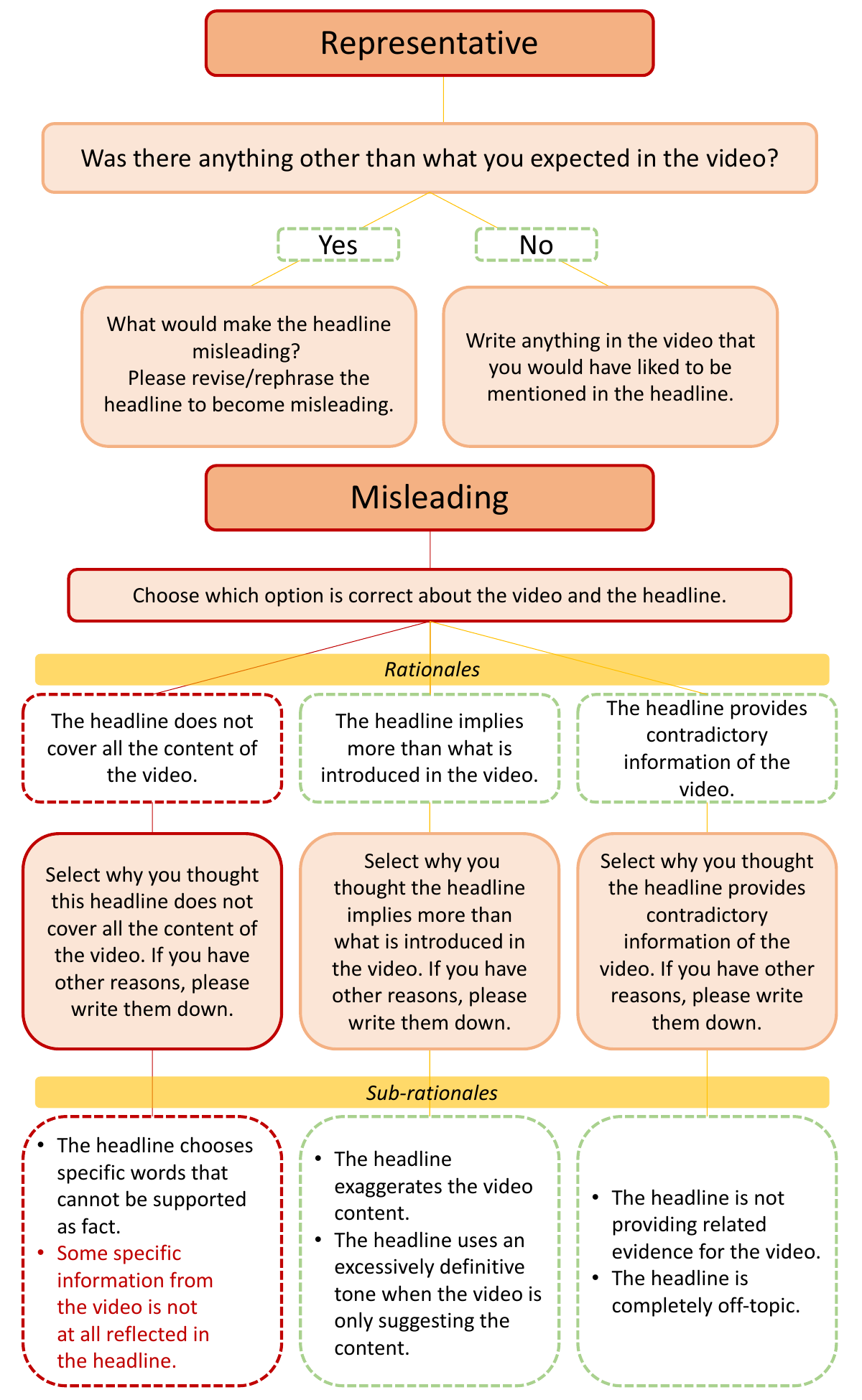}
   \caption{After label annotation, annotators provide grounding for the \emph{misleading} labels by selecting rationales and subrationales hierarchically.} 
    \label{fig:rationale}
\end{figure}

\section{Video Misleading Heading Dataset \name{}}
\label{VMH}

A \emph{misleading headline} is when the headline distorts the underlying
content~\citep{wei2017learning} and facts in the news body, leading the audience to infer more or less than what was actually presented in the content. For example, in our task, the headline ``Obama: I'm proud to be leaving \emph{without} scandal'' exaggerates the view of the content; the video plays Obama's speech that he left the administration without a \emph{significant} scandal.
This distortion makes detecting misleading video headlines even more arduous because the audience has to watch the video to know if the headline is representative or---as in this case---has a subtle exaggeration or misrepresentation.

\name{} consists of 2,247 video posts from 2014 to
2016.
We focus on this period because it coincided with the 2016 \abr{us}
presidential election, which was rife with disinformation, and is distant
enough from current events that we believe annotators can be more
confident about determining whether claims are true; as even news
organizations are not immune to false news~\citep{starbird2019disinformation}.

Our Facebook video posts come from \citet{rony2017diving},
where we manually filtered any video that exceeded five minutes or had low-quality video or sound. The videos in \name{} (Table~\ref{table:dataset_example}) average two minutes long 
and come from fifty-two media venues, including the most circulated print and broadcast media and unreliable media in the \abr{us}~\citep[listed in Appendix~\ref{venuesource} from a trustworthy journalism perspective]{edelson2021understanding, samory2020characterizing}. 

We further collect venue-related information such as venue credibility\footnote{\href{https://mediabiasfactcheck.com/}{https://mediabiasfactcheck.com/}} (e.g., High) and venue kind\footnote{\href{ https://www.pewresearch.org/journalism/fact-sheet/newspapers/audience}{https://www.pewresearch.org/journalism/fact-sheet/newspapers/audience}} (e.g., Broadcast). Also, we manually assigned news topics (e.g., Politics) inspired by News Areas\footnote{\href{https://en.wikipedia.org/wiki/News}{https://en.wikipedia.org/wiki/News}} to each headline. 
We create audio transcripts (also released in our dataset) using automatic speech recognition software\footnote{\href{https://deepgram.com}{https://deepgram.com}} whenever the video is accompanied by intelligible audio (Appendix~\ref{transcripts}). Other features in the dataset include the number of tokens per headline (average 7.75 tokens) and annotator profile (e.g., gender).  
 
\subsection{Annotation}
We ask Mechanical Turkers to identify misleading video headlines~\citep{snow-etal-2008-cheap}.
We intentionally use non-experts to reflect the world knowledge of typical web users.
For each task, the annotator goes through two phases, labeling and rationale annotation. 
We recruit three annotators per example~\citep{chandler2014nonnaivete}. 

\paragraph{Label Annotation}

 We structure the label annotation task as a series of questions to help annotators engage with the content of the headline and video (Figure~\ref{fig:diagram}).
Because headlines can take different forms (statements of facts or opinions, questions, etc.), we first ask the user to determine the form of the headline. We refer to these forms as \emph{headline property} in the rest of the paper.
Annotators get different questions depending on the headline property: if they headline is an opinion, we ask if they agree; if the headline is a fact, we ask if the think it's true (headline properties and associated questions in Appendix~\ref{Appendix:headline_property_questions}).  
This helps them build a mental model of the content of the hypothetical video \emph{before} viewing it.
We adopt this format after initial pilots indicated that directly asking if a video was misleading is too ambiguous (pilot examples in Appendix~\ref{pilotstudy}).

After the annotator has built a mental model, we ask the annotators to watch the video and answer whether the information provided in the video is consistent with the annotator's mental model of the video.
If it is, then it suggests the video is \emph{representative}: it answered the question asked by the headline, justified an opinion, or gave evidence of a new event.

In contrast, if the video fails this check, we conclude that the headline is \emph{misleading}. To
reflect the subtle difference in participants' opinions, we provide answer options that represent the levels of veracity or agreement with the headline (e.g., True, Mostly True, Mostly False, False, I don't know). For the translation to binary labels, we regard the last three answers as \emph{misleading}. 

\paragraph{Rationale Annotation} If their label is \emph{misleading}, we ask the annotators to provide a \emph{rationale}---justification---for their decision (Figure~\ref{fig:rationale}). For example, when an annotator labels a headline as \textit{misleading} and chooses \textit{The headline does not cover all the content of the video} as their rationale for the label, they then offer a subrationale to explain specifically what the headline omitted. 

We offer pre-populated rationales to force objectivity in the annotator's decision and exploit the rationales more systematically. Providing such annotations can improve not just data quality~\cite{briakou2020detecting}---by forcing the annotator to think about their reasoning---but also model accuracy~\cite{zaidan-etal-2007-using}. After the annotation is complete, final annotations are determined using a majority vote from the three annotators~\citep{yang2015wikiqa}. Because subrationales can be free-form text, we do not apply majority voting for them. 

\begin{figure}[!t]
     \centering
     \subfloat[][Qualified Workers by Accuracy Score Threshold]{\includegraphics[width=0.9
     \linewidth]{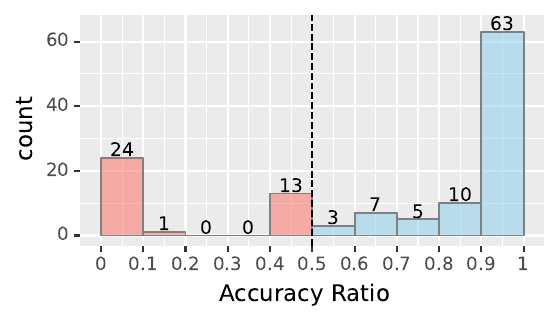}\label{figure:acc_ratio}} \\
     \subfloat[][Qualified Workers by MACE Score Threshold]{\includegraphics[width=0.9\linewidth]{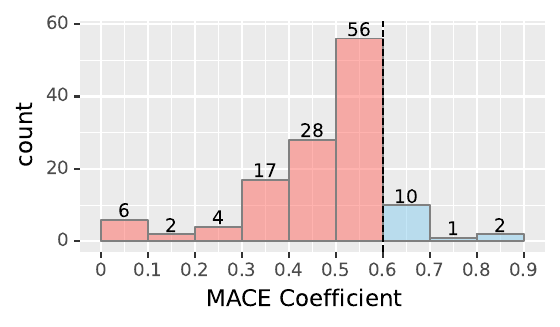}\label{figure:mace_coef}}
     \caption{\textmd{The thresholds of accuracy ratio and \abr{mace} Coefficient are manually assigned to ensure \emph{competent} workers are recruited after each annotation session.}}
     \label{qualification_threshold}
\end{figure}

\subsection{Quality Control and Assessment}
\paragraph{Quality Control}

We control the quality of \name{} to select good crowdworkers using
their accuracy score on synthetically created accuracy check questions. These
questions are synthetically created to be always misleading. For each annotator, we calculate the ratio between
the number of correct answers and the number of accuracy check
questions they answered (examples in Appendix \ref{accuracycheck}).

To determine which users are reliable and to infer the labels annotators disagree on, we use a latent variable model, \abr{mace}~\citep{paun2018comparing}, that explicitly estimates an annotator's accuracy.
This model, can correct for annotator-level biases~\citep[an annotator might overly favor a particular label, could have low overall accuracy, etc.]{martin2021crowdsourcing}. We use the point estimates---mean---from the posterior distributions of latent variables that stand for the trustworthiness of each worker (details about applying \abr{mace} to worker accuracy estimation in Appendix~\ref{Appendix:MACE}). 

As annotators enter the pool, we first vet them by asking for label annotations.
After this ``tryout'' session, annotators are reinvited only if their accuracy (0.5) or \abr{mace} score (0.6) is high enough , yielding 88 and 13 qualified workers from each metric (Figure~\ref{qualification_threshold}).

\paragraph{Quality Assessment} \label{quality_assessment}

Krippendorf's~$\alpha$ reveals the difficulty of the task and the quality of the annotators: for the three annotators who passed the accuracy score threshold, it was 0.57 for labels and 0.33 for rationales.
The Krippendorf’s~$\alpha$ values of the workers who qualified with the \abr{mace} cutoff are 0.68 (labels) and 0.21 (rationales).
While the values have moderate-to-low agreement \citep{briakou2020detecting}, this is expected due to the inherent subjectivity of the annotation~\citep{sandri-etal-2023-dont, kenyon-dean-etal-2018-sentiment,akhtar2019new, daume2005bayesian}.
These inevitable disagreements are important as they can help capture the task's nuance~\citep{davani2022dealing}: the \emph{source} of the disagreements can be revealing, as we discuss more in the next section.
\section{Dataset Analysis}\label{section3}

\begin{figure}[!t]
     \centering
     \subfloat[][Venue Distribution]{\includegraphics[width=0.98
     \linewidth]{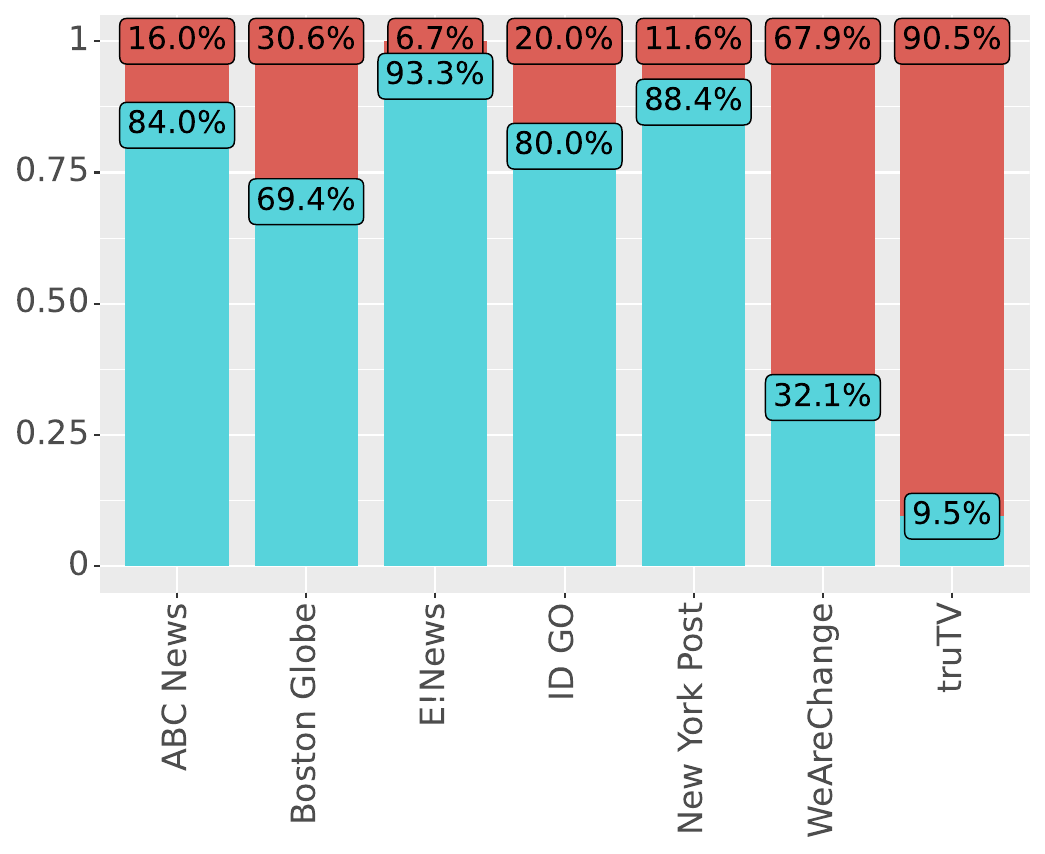}\label{figure:venue}} \\
     \subfloat[][Venue Kind Distribution]{\includegraphics[width=0.98\linewidth]{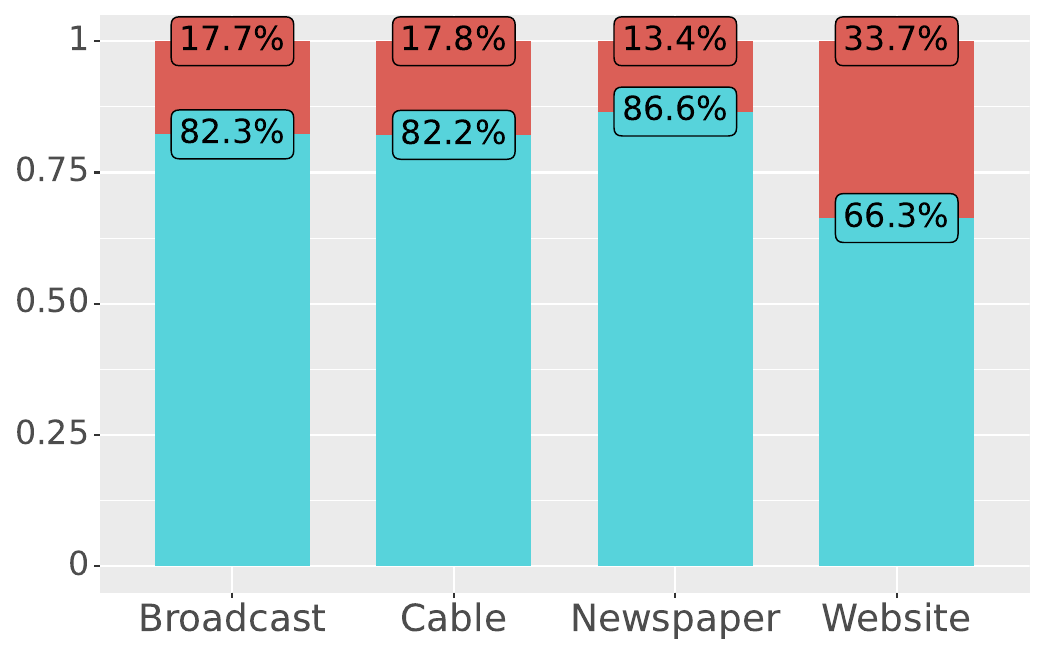}\label{figure:venue_kind}}
     \caption{What proportion of headlines were misleading (red) or
       representative (blue) based on
       specific venues (top) and venue types (bottom).
       \textmd{The venues \emph{TruTV}, \emph{WeAreChange.org} and
         venue kind \emph{Website} were the strongest indicators of
         misleading headlines.
         The red and blue bars are proportions of \emph{misleading} and \emph{representative} labels.
         Not all venues are shown.}
       }
     \label{fig:distribution}
\end{figure}

Out of 2,247 video headlines, 1,906 headlines are annotated as
\emph{\representative}, while 341 headlines are
annotated as \emph{misleading}, suggesting a high class
imbalance. This section investigates \name{} to understand what
features contribute to (or correlate with) a headline being classified
as misleading.

\paragraph{Misleading Features} Figure~\ref{fig:distribution} suggests
that the venues \emph{TruTV} and \emph{WeAreChange.org} are strong
indicators for misleading headlines. More generally, videos from the
venue kind \emph{Website} (as opposed to traditional media) are likely
to be misleading (29\%).
The specific venue and the kind of venue may help detect misleading headlines (Appendix~\ref{metadata}).

\paragraph{Clickbait} Misleading videos and clickbait both have the same goal: to entice more people to click on the underlying content.  
A reasonable hypothesis is that they would use similar tricks to lure in users. 
Thus, we reproduce the features found by \citep{dhoju2019differences} to be associated with clickbait headlines such as the number of demonstrative adjectives, numbers, and WH-words (e.g., what, who, how) for the headlines in \name{}.
Demonstrative adjectives do appear more frequently in misleading
headlines, while numbers and superlative word features are less
frequent (Table~\ref{tab:clickbait_patterns}). Numbers and modal words
appear in similar frequencies. Thus, misleading video headlines are
not the same as clickbait.

\begin{table}[!t]
    \centering
    \scalebox{0.8}{
    \begin{tabular}{cccc}
    \Xhline{1.5pt}
    \rowcolor{Gray}
    & \multicolumn{2}{c}{\textbf{Presence Ratio}} \\
    \rowcolor{Gray}
          \multirow{-2}{*}{\textbf{Clickbait Patterns}}     &\citet{dhoju2019differences} & \textbf{VMH (Ours)}
         \\ 
         \Xhline{1.5pt}
         {Demonstrative  Adj}    & 0.80 & 0.61 \\
         WH-Words                & 0.70 & 0.40 \\
         Numbers                 & 0.72 & 0.60 \\
         Modal                   & 0.27 & 0.20 \\
         Superlative             & 0.30 & 0.06 \\
    \Xhline{1.5pt}
    \end{tabular}}
    \caption{Clickbait patterns in misleading headlines in \name{} to demonstrate the difference between clickbait detection and misleading video headline task.}
    \label{tab:clickbait_patterns}
\end{table}

\paragraph{Investigation of Bias in Annotation}
Because our dataset has many politically relevant videos, we also ask
annotators' political leaning to see if it biases annotations.
A $\chi^2$ test does not suggest that annotations and political
leanings are dependent ($p$-value 0.36); the marginal proportion of
misleading videos are comparable (Democratic: 22.9\%, Republican:
22.6\%, and Independent: 33\%).

We also manually check fifty video headlines to see if their
ideologies affected a headline’s assigned label, finding no
substantial consequences. For example, the headline ``Charles Blow:
Donald Trump is a bigot'', presumably ``anti-Trump'', was annotated
\emph{Representative} by an annotator with a ``Republican'' leaning.

\begin{table*}[t]
    \renewcommand{\arraystretch}{1.01}
    \small
    \centering
    \setlength{\tabcolsep}{4pt}
    \scalebox{0.9}{\begin{tabular}{ccccc}
    \Xhline{1.5pt}
    \rowcolor{Gray}
         \textbf{Headlines} & \textbf{ID} & \textbf{Ann.} & \textbf{Rationales} & \textbf{Subrationales} \\[1.1ex] \Xhline{1.5pt}
         & \multirow{2}{*}{81} & \multirow{2}{*}{M} &  The headline does not cover all the & The headline is not providing related\\
                  &                    &                             &  content of the video & evidence for the video \\
         Lester Holt Interrupted  & \multirow{2}{*}{111} & \multirow{2}{*}{M} &  Neither of above: The headline provides & The headline chooses specific words \\
         Trump Repeatedly         &                    &                             & contradictory information of the video & that cannot be supported as fact \\
         & \multirow{2}{*}{97} & \multirow{2}{*}{R} & \multirow{2}{*}{-} & \multirow{2}{*}{-} \\ 
         &&&&\\ \cdashline{1-5}
         & \multirow{2}{*}{42} & \multirow{2}{*}{M} &  The headline does not cover all the & The headline chooses specific words\\
         Emily Blunt Weighs In          &                    &                             &  content of the video & that cannot be supported as fact \\
         On John Kransinskis& \multirow{2}{*}{45} & \multirow{2}{*}{M} &  The headline does not cover all the & Some specific information from the  \\
         Obsession With The         &                    &                             &  content of the video & video  is not at all reflected in the headline \\
          D...& \multirow{2}{*}{97} & \multirow{2}{*}{R} & \multirow{2}{*}{-} & \multirow{2}{*}{-} \\ 
          &&&&\\ \cdashline{1-5}
          & \multirow{2}{*}{77} & \multirow{2}{*}{M} &  Neither of above: The headline provides & The headline is not providing related\\
          &                    &                             & contradictory information of the video & evidence for the video \\
          Did This Man Murder & \multirow{3}{*}{12} & \multirow{3}{*}{M} &  \multirow{2}{*}{The headline implies more than what} & The headline uses an excessively  \\
          A Beautiful Country &                    &                             & \multirow{2}{*}{what is introduced in the video} & definitive  tone when the video is   \\
          Music Producer &&&& only suggesting the content \\
           & \multirow{2}{*}{10} & \multirow{2}{*}{M} &  Neither of above: The headline provides & (Free Form Input) No mention of her   \\
          &                    &                             & contradictory information of the video & being a country music producer \\ \Xhline{1.5pt}
    \end{tabular}}
    \caption{Examples of Samples with Subjectivity. The second headline shows that each annotator's rationales are different even when the annotations are the same. The third headline shows an example where annotated subrationales all vary in their content (e.g., free-form text). ID is Annotator's ID and Ann.\ is the annotation result from each annotator (M: Misleading, R: Representative) }
    \label{tab:rationales}
\end{table*}

\paragraph{Task Subjectivity} \label{task subjectivity}
Motivated by Section~\ref{quality_assessment}, we examine the annotations that fail to have consensus among annotator decisions: there were 1436 \emph{\representative} and 159 \emph{misleading} instances with the perfect agreement, leaving $30$\% to annotations that had disagreement. 
In addition to disagreeing on labels, annotators disagree about why the headline is misleading (Table~\ref{tab:rationales}).

\section{Experiments} \label{Experiments}

The misleading headline detection task is challenging because of the inherent subjectivity of the task.
It also requires multimodal approaches that can consider both the headline and the video to make inferences about whether the headline is \emph{representative} or not.
Thus, this section benchmarks both text-only and multimodal approaches typically used for detecting video-text similarity and video-text entailment tasks. 

\paragraph{Experiment Settings} 
We compare the performance of models when trained with various combinations of input features in our dataset. The features that we consider are headlines ($H$) and their associated video clips ($V$), transcripts ($T$), rationales, and sub-rationales ($R$). 

For textual features, we concatenate features as:\footnote{While gold rationales might not be available during inference, our objective to study them as features are to highlight and understand if and how rationales can help improve detection accuracy in this task. We leave automatic prediction of the rationales to future work. } \texttt{[SEP]} \{Headline \texttt{[SEP]} Transcript \texttt{[SEP]} rationale \texttt{[SEP]} sub-rationale\}.
We also extract embeddings corresponding to two multimodal models. We use VideoCLIP \cite{xu2021videoclip} and \abr{vlm} models  \cite{xu2021vlm} that adopt zero-shot transfer learning to video-text understanding tasks\footnote{The benchmark results in our study are to suggest baseline features and models that could be used in solving the detection task, rather than demonstrating them as a sole approach to validate the dataset or improve the detection performance. }.
VideoCLIP trains a transformer model using a contrastive objective on paired examples of video-text clips that maximize association between temporarily overlapping text and video segments \citep{xu2021videoclip}. In contrast, \abr{vlm} is a task-agnostic multimodal learning model that uses novel masking schemes to improve the learning of multimodal fusion between the text and the video. 
We finetune a classification layer that takes input features extracted from video and text-based encoders as described to predict the label associated with a given video-headline pair (details in Appendix \ref{finetune}).
\paragraph{Data and Evaluation Metrics}
We divide \name{} into three sets: 70\% for the training set, 15\% for the validation set, and 15\% for the test set. We evaluate using the following metrics: F1, precision, recall, \abr{auprc} score, and accuracy. We report the precision and recall scores of the positive class, \emph{misleading}. Each metric is estimated by averaging five replicates of stratified random splits.

\section{Experimental Results and Model Analyses} 
\begin{table*}[t]
    \small
    \centering
    \setlength{\tabcolsep}{8pt}
    \renewcommand{\arraystretch}{1.2}
    \begin{tabular}{clccccc}
        
        \Xhline{1.5pt}
        \rowcolor{Gray}
        
         &  & \multicolumn{5}{c}{\textbf{Evaluation Metrics}} \\ 
        \rowcolor{Gray}
        
        \multirow{-2}{*}{\textbf{Model}}&\multicolumn{1}{c}{\multirow{-2}{*}{\textbf{Input}}} & F1-Score & Precision & Recall & AUPRC & Accuracy  \\ \Xhline{1.5pt}
        
        \multirow{3}{*}{\textbf{BERT}}  
                                & \textbf{H}
                                & \colorbox{green!15}{\textbf{0.16 (0.07)}} & \textbf{0.29 (0.14)} & 0.11 (0.05) & \textbf{0.17 (0.02)} &  \textbf{0.82 (0.01)} \\
                               & \textbf{H + T} 
                                & 0.16 (0.08) & 0.26 (0.11) & \textbf{0.12 (0.06)} & 0.15 (0.01) &  \textbf{0.82 (0.01)}\\
        \hline
        \multirow{5}{*}{\textbf{VideoCLIP}}  
                               & \textbf{H} 
                                & 0.16 (0.06) & 0.22 (0.05) & 0.13 (0.06) & 0.17 (0.01) &  0.80 (0.01)\\
                               & \textbf{V} 
                               & 0.17 (0.03) & 0.25 (0.06) & 0.14 (0.04) & 0.16 (0.00) &  0.79 (0.02)\\ 
                               & \textbf{V + H} 
                                & 0.26 (0.09) & 0.32 (0.13) & 0.24 (0.09) & 0.20 (0.04) &  0.79 (0.05) \\ 
                               & \textbf{V + H + T} 
                                & 0.21 (0.04) & 0.29 (0.06) & 0.17 (0.03) & 0.17 (0.01) &  0.80 (0.01)\\
                               & \textbf{V + H + T + R}
                                & \colorbox{green!15}{\textbf{0.53 (0.06)}} & \textbf{0.65 (0.08)} & \textbf{0.44 (0.06)} & \textbf{0.41 (0.05)} & \textbf{0.88 (0.01)} \\

        \hline
        \multirow{5}{*}{\textbf{VLM}}  
                           & \textbf{H} 
                            & 0.18 (0.05) & 0.20 (0.06) & 0.19 (0.09) & 0.16 (0.01) &  0.76 (0.04) \\
                           & \textbf{V} 
                            & 0.00 (0.00) & 0.00 (0.00) & 0.00 (0.00) & 0.15 (0.00) &  0.83 (0.00) \\
                           & \textbf{V + H} 
                           & 0.22 (0.06) & 0.23 (0.05) & 0.22 (0.06) & 0.18 (0.02) &  0.77 (0.02) \\
                           & \textbf{V + H + T} 
                            & 0.23 (0.04) & 0.23 (0.04) & 0.56 (0.01) &  0.17 (0.01) & 0.76 (0.01)  \\
                           & \textbf{V + H + T + R}
                            & \colorbox{green!15}{\textbf{0.56 (0.03)}} & \textbf{0.63 (0.02)} & \textbf{0.52 (0.05)} & \textbf{0.40 (0.03)} &  \textbf{0.88 (0.00)}\\ \Xhline{1.5pt}
    \end{tabular}
    \caption{Benchmark Evaluation Results. Rows for each model shows performance with different input features: headlines (\textbf{H}), videos (\textbf{V}), transcripts (\textbf{T}), and rationales (\textbf{R}). The reported metrics are the average F1-score, average Precision score, average Recall score, average AUPRC score, and average accuracy score of 5 replicates of stratified random splits of the train, valid, and test sets. The brackets indicate standard deviation for each metric.}
    \label{table:experiment_result}
\end{table*}
\paragraph{Experiment Results} Table~\ref{table:experiment_result} reports the main results: the multimodal models that use all the features, \{Video Frame + Headline + Transcript + Rationale (V+H+T+R)\} result in the best performance across the board, outperforming text-only based model.
Adding rationales obviously helps, as these were the foundation of the annotator labels, and \textit{subrationales} help even more (Appendix~\ref{subrationales}).

Next, we validate the utility of the multimodal features in a partial-input setting.
We explore how the subjectivity can affect the detection.
%

\paragraph{Partial Input Analysis}
Validating a dataset with a partial-input baseline is common in multimodal datasets~\citep{thomason2019single}. Artifacts in the dataset can lead the models to \emph{cheat} using shortcut features that can result in poor generalizability \citep{feng2019misleading}. Thus, in our case, we also experiment with unimodal settings (partial input)---\{Video\} and \{Headline\}---to ensure \name{} does not contain such artifacts. Using only video or text-based features result in poor F1 ($0.16-0.18$) relative to multimodal features (F1-score: > $0.22$). 

\paragraph{Model Subjectivity Analysis} To understand the subjectivity of the task (Section~\ref{task subjectivity}), we also report F1-scores on the subset of the dataset, \textit{subjective} samples ($30$\%), that had low consensus in the annotation process. Training on this subset, even the best model with all features: \{Video from VideoCLIP + Headline + Transcript + Rationale\} only obtains 0.12 F1; and it drops to 0.10 with \abr{vlm} compared to 0.53 (VideoCLIP) and 0.56 \abr{vlm} using the entire training set.
Difficult instances for humans might not include any reliable features for the model.
 
\paragraph{Video-Text Entailment Analysis}
A sceptical reader might content that this task problem is just entailment: if the headline is entailed from the video, it is representative.
However, this is not a complete solution: to investigate the relationship we use transcripts to stand in for the video and then ask the RoBERTa \abr{nli} model\footnote{\href{https://huggingface.co/ynie/roberta-large-snli_mnli_fever_anli_R1_R2_R3-nli}{fine-tuned on SNLI, MNLI, FEVER-NLI, and ANLI}} whether the headline is entailed from the transcript.
We average the entailment score between chunked sentences from transcripts and the headlines to compensate for different lengths.
To calculate if there is correlation between entailment predictions and the labels, we conduct a $t$-test~\citep{gerald2018brief}.
The $p$-value is 0.01, which indicates that the difference between the two is statistically significant: this is a signal. 

However, it is not a stand-alone solution; Table~\ref{tab:entailment} shows examples when entailment decisions contradict the annotator's judgments.
For example, the first headline shows a high entailment score with the transcript while annotated as misleading with the rationale of ``The headline does not cover all the video content''. The second and third headlines are predicted with low entailment scores or ``not entail'' while being annotated as \emph{representative} by majority annotators. 

\begin{table*}[!t]
\centering
    \small
    \resizebox{\textwidth}{!}{
    \scalebox{0.9}{\begin{tabular}{p{25mm} p{80mm} ccc}
    \Xhline{1.5pt}
    \rowcolor{Gray}
     \textbf{Headlines} & \textbf{Transcripts} & \textbf{Entail} & \textbf{Score} & \textbf{Answer} \\[1.1ex] \Xhline{1.5pt}
     The sounds of emotions & \dots We use the principles of music to work with rhythm and melody to regain the functional use of language. Phrase is if we\dots Nice job. Let's all. Well You wanna skip this up? Okay. Do you wanna skip it or singing it? You wanna try to sing it? Let's jump to the chorus. Okay? So darling then. Music is what emotions sound like\dots & \ding{51} & 0.71 & M \\
     There is a double standard & \dots Is there a double standard when it comes to transparency between Trump and Clinton? Well, of course, there's a double standard\dots He's doing over a hundred foreign deals and he wants to be both the commander chief and the representative in the world for the United States\dots I mean, the difference between telling somebody you had pneumonia on Sunday instead of Friday is not even in the same league really\dots& \ding{55} & 0.20 & R \\
     Honor a Vet I Warfighters & \dots Having worked with veterans throughout my career, I know firsthand the importance of honoring our troops. This veterans day our series the war fighters and history are partnering with Team Rub con to create honor event\dots Honor the vets and more fighters in your life, and share a photo and a story today. Learn more history dot com honor that\dots & \ding{51}  & 0.53 & R \\ \Xhline{1.5pt}
    \end{tabular}}}
    \caption{Examples that show entailment is not enough to discover misleading headlines. The first headline shows high entailment score with the transcript while annotated as \emph{misleading} with the rationale of ``The headline does not cover all the content of the video''. The second and third headline are predicted with low entailment score or ``not entail'' while being annotated as ``representative'' by majority annotators.}
    \label{tab:entailment}
\end{table*}

\section{Related Work} \label{related}
One of the major factors of misinformation is inaccurate headlines, which pervade social media platforms\citep{wei2017learning}. Clickbait is characterized by misleading headlines, depending on the degree of deception the audience experiences \citep{bourgonje2017clickbait}. However, clickbait detection problems are distinguished from misleading headlines as they may exaggerate the content but are not particularly misleading \citep{chen2015misleading}. 

As the spread of fake news appears in many forms of multimedia \citep{aimeur2023fake}, several works are on constructing datasets to enable research on multimodal misleading headline detection \citep{bu2023online}. 
\citet{ha2018characterizing} introduces an image-based dataset and focuses on misrepresented headlines on Instagram. 
Also, \citet{shang2019towards} introduces a dataset of Youtube videos with manual annotations generated by misleading seed videos from the Youtube recommendation system. 
\citet{zannettou2018good} proposes a misleading-labeling mechanism with both manual and automatic. In this case, annotated videos could be biased as manual and automatic annotation may not be in consensus; they can lead to erroneous annotations of misleading headlines.

Apart from dataset research, previous works focus on detecting multimodal fake news by including multimedia features such as false videos, images, audio, and caption \citep{fakesv, masciari2020detecting, demuyakor2022fake, mccrae2022multi}. However, these works feature general forms of fake news (i.e., deep-fake videos), not misleading headlines. 

For multimodal models built for misleading headline detection tasks, \citet{song2016click} identified the video thumbnails, \citet{li2022cnn} uses uploader and environment features (e.g., number of likes received, the date of most recent upload), \citet{choi2022effective} uses comments and domain knowledge, and \citet{zannettou2018good} uses video's meta statistics (e.g., number of shares) to develop a deep variational autoencoder with semi-supervised learning. \citet{shang2019towards} uses a convolutional neural network approach to find the correlation between the neural net features and the headline. \citet{you2023video} uses model-selected video frames as input features to the classifier to detect dissimilarity between the video and the text. 

\section{Conclusion and Future Work}

We present \name{}, a dataset of misleading headlines from social media videos.
Our annotation scheme reduces the task's subjectivity, and we verify the
reliability of the annotations.
We believe incorporating the crowd workers' distinct opinions (e.g., headline types and rationales) on misleading headlines allows crude reflection of the current social media misinformation phenomenon.
Through their lenses, we anticipate a better understanding of how people perceive misinformation in misleading video headlines and for future work, use it to generalize the detection models that are soon to be deployed. 

To obtain even more realistic examples for this task, we encourage applying a dynamic adversarial generation pipeline. Motivated by \citet{eisenschlos2021fool}, misleading headlines could be authored by humans guided to break the existing video headline detection models. For example, while they are writing a \textit{misleading} headline, if the model falsely predicts the headline as \textit{representative}, it would become an adversarial, \textit{realistic} example~\citep{ma2021dynaboard}. These examples can prevent the model from learning superficial patterns \citep{kiela-etal-2021-dynabench} and further be developed to become a \textit{robust} tool for journalists to prevent them from making ``honest'' mistakes when writing video headlines~\citep{Dhiman}.

\section{Limitations}
 Although the rationales advance the model's knowledge in detecting misleading headlines, the limitation of this paper is that gold rationales are not realistic. Thus, the current rationale setting can be set as an upper bound for the generic model evaluation. Also, when building the model, we suggest including features that are alike with ``subrationale'' features in \name{}, which informs \textit{how} a headline is misleading. 

Moreover, we acknowledge that the visual grounding of the headline may help the model to learn how the headline is (partially) relevant to the video’s visual content. It would be interesting to see what other multimodal models with visual grounding ability could be applied to our task; a multimodal model could be designed so that it addresses the questions of whether the headline represents the message the video conveys or identifying the gap between the video message and the headline.

\section{Ethical Considerations} \label{Ethics}
We address ethical considerations for dataset papers, given that our work proposes a new dataset \name{}. We reply to the relevant questions posed in the {\texttt{\abr{acl} 2022 Ethics \abr{faq}}}\footnote{https://www.acm.org/code-of-ethics}. 

To collect \name{} videos, we follow the community guidelines by Meta by using publicly available videos that are accessible with \emph{public-view only} accounts. Our study was pre-monitored by an official \abr{irb} review board to protect the participants' privacy rights. Moreover, the identity characteristics of the participants were self-identified by the workers by answering the survey questions. 

Before distributing the survey, we collected consent forms for the workers to agree that their answers would be used for academic purposes. All workers who make good faith annotations are paid regardless of their accuracy.
The MTurkers were compensated over $10$ \abr{usd} an hour (a rate higher than the \abr{us} national minimum wage of $7.50$ \abr{usd}
). 

Although we understand that \name{} may be exploited to make misleading content in the future, we emphasize the impact of its social goods; it provides the resource to combat multimodal misinformation online today. As \name{} is the first dataset that introduces video for misleading headline detection, we believe it will serve as a starting point in the research community to overcome the task. 

\paragraph{Acknowledgements}
We thank CLIP and CJ Lab members and the anonymous reviewers for their insightful feedback. We thank the user study participants for supporting this work through annotating data. Yoo Yeon Sung, Naeemul Hassan, and Jordan Boyd-Graber are supported in part by NSF Grant ``BaitBuster 2.0: Keeping Users Away From Clickbait'' and DARPA Grant ``SHADE'' projects. 
\bibliography{bib/yy}

\begin{thebibliography}{54}
\expandafter\ifx\csname natexlab\endcsname\relax\def\natexlab#1{#1}\fi

\bibitem[{A{\"\i}meur et~al.(2023)A{\"\i}meur, Amri, and
  Brassard}]{aimeur2023fake}
Esma A{\"\i}meur, Sabrine Amri, and Gilles Brassard. 2023.
\newblock \href {https://doi.org/10.1007/s13278-023-01028-5} {Fake news,
  disinformation and misinformation in social media: a review}.
\newblock \emph{Social Network Analysis and Mining}, 13(1):30.

\bibitem[{Akhtar et~al.(2019)Akhtar, Basile, and Patti}]{akhtar2019new}
Sohail Akhtar, Valerio Basile, and Viviana Patti. 2019.
\newblock A new measure of polarization in the annotation of hate speech.
\newblock In \emph{AI* IA 2019--Advances in Artificial Intelligence: XVIIIth
  International Conference of the Italian Association for Artificial
  Intelligence, Rende, Italy, November 19--22, 2019, Proceedings 18}, pages
  588--603. Springer.

\bibitem[{Allcott et~al.(2019)Allcott, Gentzkow, and Yu}]{allcott2019trends}
Hunt Allcott, Matthew Gentzkow, and Chuan Yu. 2019.
\newblock \href {https://journals.sagepub.com/doi/pdf/10.1177/2053168019848554}
  {Trends in the diffusion of misinformation on social media}.
\newblock \emph{Research \& Politics}, 6(2):2053168019848554.

\bibitem[{Bourgonje et~al.(2017)Bourgonje, Schneider, and
  Rehm}]{bourgonje2017clickbait}
Peter Bourgonje, Julian~Moreno Schneider, and Georg Rehm. 2017.
\newblock \href {https://aclanthology.org/W17-4215} {From clickbait to fake
  news detection: an approach based on detecting the stance of headlines to
  articles}.
\newblock In \emph{Proceedings of the 2017 EMNLP workshop: natural language
  processing meets journalism}, pages 84--89.

\bibitem[{Briakou and Carpuat(2020)}]{briakou2020detecting}
Eleftheria Briakou and Marine Carpuat. 2020.
\newblock Detecting fine-grained cross-lingual semantic divergences without
  supervision by learning to rank.
\newblock In \emph{Proceedings of the 2020 Conference on Empirical Methods in
  Natural Language Processing (EMNLP)}, pages 1563--1580.

\bibitem[{Bu et~al.(2023)Bu, Sheng, Cao, Qi, Wang, and Li}]{bu2023online}
Yuyan Bu, Qiang Sheng, Juan Cao, Peng Qi, Danding Wang, and Jintao Li. 2023.
\newblock \href
  {https://ui.adsabs.harvard.edu/abs/2023arXiv230203242B/abstract} {Online
  misinformation video detection: A survey}.
\newblock \emph{arXiv e-prints}, pages arXiv--2302.

\bibitem[{Chandler et~al.(2014)Chandler, Mueller, and
  Paolacci}]{chandler2014nonnaivete}
Jesse Chandler, Pam Mueller, and Gabriele Paolacci. 2014.
\newblock \href {https://link.springer.com/article/10.3758/s13428-013-0365-7}
  {Nonna{\"\i}vet{\'e} among amazon mechanical turk workers: Consequences and
  solutions for behavioral researchers}.
\newblock \emph{Behavior research methods}, 46:112--130.

\bibitem[{Chen et~al.(2015)Chen, Conroy, and Rubin}]{chen2015misleading}
Yimin Chen, Niall~J Conroy, and Victoria~L Rubin. 2015.
\newblock \href {https://doi.org/10.1145/2823465.2823467} {Misleading online
  content: recognizing clickbait as" false news"}.
\newblock In \emph{Proceedings of the 2015 ACM on workshop on multimodal
  deception detection}, pages 15--19.

\bibitem[{Chesney et~al.(2017)Chesney, Liakata, Poesio, and
  Purver}]{chesney2017incongruent}
Sophie Chesney, Maria Liakata, Massimo Poesio, and Matthew Purver. 2017.
\newblock \href {https://aclanthology.org/W17-4210} {Incongruent headlines: Yet
  another way to mislead your readers}.
\newblock In \emph{Proceedings of the 2017 EMNLP Workshop: Natural Language
  Processing meets Journalism}, pages 56--61.

\bibitem[{Choi and Ko(2022)}]{choi2022effective}
Hyewon Choi and Youngjoong Ko. 2022.
\newblock \href
  {https://www.sciencedirect.com/science/article/pii/S0167865522000071}
  {Effective fake news video detection using domain knowledge and multimodal
  data fusion on youtube}.
\newblock \emph{Pattern Recognition Letters}, 154:44--52.

\bibitem[{Daume~III and Marcu(2005)}]{daume2005bayesian}
Hal Daume~III and Daniel Marcu. 2005.
\newblock \href
  {https://www-nlpir.nist.gov/projects/duc/pubs/2005papers/isi.daume.pdf}
  {Bayesian summarization at duc and a suggestion for extrinsic evaluation}.
\newblock In \emph{Proceedings of the Document Understanding Conference,
  DUC-2005, Vancouver, USA}.

\bibitem[{Davani et~al.(2022)Davani, D{\'\i}az, and
  Prabhakaran}]{davani2022dealing}
Aida~Mostafazadeh Davani, Mark D{\'\i}az, and Vinodkumar Prabhakaran. 2022.
\newblock Dealing with disagreements: Looking beyond the majority vote in
  subjective annotations.
\newblock \emph{Transactions of the Association for Computational Linguistics},
  10:92--110.

\bibitem[{Demuyakor and Opata(2022)}]{demuyakor2022fake}
John Demuyakor and Edward~Martey Opata. 2022.
\newblock \href {https://doi.org/10.54963/jic.v2i1.56} {Fake news on social
  media: Predicting which media format influences fake news most on facebook}.
\newblock \emph{Journal of Intelligent Communication}, 2(1).

\bibitem[{Dhiman(2023)}]{Dhiman}
Bharat Dhiman. 2023.
\newblock \href {https://papers.ssrn.com/sol3/papers.cfm?abstract_id=4401194}
  {Does artificial intelligence help journalists: A boon or bane?}

\bibitem[{Dhoju et~al.(2019)Dhoju, Main Uddin~Rony, Ashad~Kabir, and
  Hassan}]{dhoju2019differences}
Sameer Dhoju, Md~Main Uddin~Rony, Muhammad Ashad~Kabir, and Naeemul Hassan.
  2019.
\newblock \href {https://doi.org/10.1145/3308560.3316741} {Differences in
  health news from reliable and unreliable media}.
\newblock In \emph{Companion Proceedings of The 2019 World Wide Web
  Conference}, pages 981--987.

\bibitem[{dos Rieis et~al.(2015)dos Rieis, de~Souza, de~Melo, Prates, Kwak, and
  An}]{dos2015breaking}
Julio Cesar~Soares dos Rieis, Fabr{\'\i}cio~Benevenuto de~Souza, Pedro Olmo
  S~Vaz de~Melo, Raquel~Oliveira Prates, Haewoon Kwak, and Jisun An. 2015.
\newblock \href {https://doi.org/10.1609/icwsm.v9i1.14619} {Breaking the news:
  First impressions matter on online news}.
\newblock In \emph{Ninth International AAAI Conference on Web and Social
  Media}.

\bibitem[{Edelson et~al.(2021)Edelson, Nguyen, Goldstein, Goga, McCoy, and
  Lauinger}]{edelson2021understanding}
Laura Edelson, Minh-Kha Nguyen, Ian Goldstein, Oana Goga, Damon McCoy, and
  Tobias Lauinger. 2021.
\newblock \href {https://doi.org/10.1145/3487552.3487859} {Understanding
  engagement with us (mis) information news sources on facebook}.
\newblock In \emph{Proceedings of the 21st ACM Internet Measurement
  Conference}, pages 444--463.

\bibitem[{Eisenschlos et~al.(2021)Eisenschlos, Dhingra, Bulian,
  B{\"o}rschinger, and Boyd-Graber}]{eisenschlos2021fool}
Julian Eisenschlos, Bhuwan Dhingra, Jannis Bulian, Benjamin B{\"o}rschinger,
  and Jordan Boyd-Graber. 2021.
\newblock \href {https://aclanthology.org/2021.naacl-main.32} {Fool me twice:
  Entailment from wikipedia gamification}.
\newblock In \emph{Proceedings of the 2021 Conference of the North American
  Chapter of the Association for Computational Linguistics: Human Language
  Technologies}, pages 352--365.

\bibitem[{Feng et~al.(2019)Feng, Wallace, and Boyd-Graber}]{feng2019misleading}
Shi Feng, Eric Wallace, and Jordan Boyd-Graber. 2019.
\newblock \href {https://aclanthology.org/P19-1554} {Misleading failures of
  partial-input baselines}.
\newblock In \emph{Proceedings of the 57th Annual Meeting of the Association
  for Computational Linguistics}, pages 5533--5538.

\bibitem[{Gerald(2018)}]{gerald2018brief}
Banda Gerald. 2018.
\newblock \href
  {https://www.sciencepublishinggroup.com/journal/paperinfo?journalid=322&paperId=10031643}
  {A brief review of independent, dependent and one sample t-test}.
\newblock \emph{International journal of applied mathematics and theoretical
  physics}, 4(2):50--54.

\bibitem[{Ha et~al.(2018)Ha, Kim, Won, Cha, and Joo}]{ha2018characterizing}
Yui Ha, Jeongmin Kim, Donghyeon Won, Meeyoung Cha, and Jungseock Joo. 2018.
\newblock \href {https://doi.org/10.1609/icwsm.v12i1.15019} {Characterizing
  clickbaits on instagram}.
\newblock In \emph{Proceedings of the International AAAI Conference on Web and
  Social Media}, volume~12.

\bibitem[{Hovy et~al.(2013)Hovy, Berg-Kirkpatrick, Vaswani, and
  Hovy}]{hovy2013learning}
Dirk Hovy, Taylor Berg-Kirkpatrick, Ashish Vaswani, and Eduard Hovy. 2013.
\newblock \href {https://aclanthology.org/N13-1132} {Learning whom to trust
  with mace}.
\newblock In \emph{Proceedings of the 2013 Conference of the North American
  Chapter of the Association for Computational Linguistics: Human Language
  Technologies}, pages 1120--1130.

\bibitem[{Kenyon-Dean et~al.(2018)Kenyon-Dean, Ahmed, Fujimoto,
  Georges-Filteau, Glasz, Kaur, Lalande, Bhanderi, Belfer, Kanagasabai,
  Sarrazingendron, Verma, and Ruths}]{kenyon-dean-etal-2018-sentiment}
Kian Kenyon-Dean, Eisha Ahmed, Scott Fujimoto, Jeremy Georges-Filteau,
  Christopher Glasz, Barleen Kaur, Auguste Lalande, Shruti Bhanderi, Robert
  Belfer, Nirmal Kanagasabai, Roman Sarrazingendron, Rohit Verma, and Derek
  Ruths. 2018.
\newblock \href {https://doi.org/10.18653/v1/N18-1171} {Sentiment analysis:
  It{'}s complicated!}
\newblock In \emph{Proceedings of the 2018 Conference of the North {A}merican
  Chapter of the Association for Computational Linguistics: Human Language
  Technologies, Volume 1 (Long Papers)}, pages 1886--1895, New Orleans,
  Louisiana. Association for Computational Linguistics.

\bibitem[{Kiela et~al.(2021)Kiela, Bartolo, Nie, Kaushik, Geiger, Wu, Vidgen,
  Prasad, Singh, Ringshia, Ma, Thrush, Riedel, Waseem, Stenetorp, Jia, Bansal,
  Potts, and Williams}]{kiela-etal-2021-dynabench}
Douwe Kiela, Max Bartolo, Yixin Nie, Divyansh Kaushik, Atticus Geiger,
  Zhengxuan Wu, Bertie Vidgen, Grusha Prasad, Amanpreet Singh, Pratik Ringshia,
  Zhiyi Ma, Tristan Thrush, Sebastian Riedel, Zeerak Waseem, Pontus Stenetorp,
  Robin Jia, Mohit Bansal, Christopher Potts, and Adina Williams. 2021.
\newblock \href {https://doi.org/10.18653/v1/2021.naacl-main.324} {Dynabench:
  Rethinking benchmarking in {NLP}}.
\newblock In \emph{Proceedings of the 2021 Conference of the North American
  Chapter of the Association for Computational Linguistics: Human Language
  Technologies}, pages 4110--4124, Online. Association for Computational
  Linguistics.

\bibitem[{Li et~al.(2022)Li, Xiao, Li, Hu, Yao, and Li}]{li2022cnn}
Xiaojun Li, Xvhao Xiao, Jia Li, Changhua Hu, Junping Yao, and Shaochen Li.
  2022.
\newblock \href {https://europepmc.org/article/med/35414095} {A cnn-based
  misleading video detection model}.
\newblock \emph{Scientific Reports}, 12(1):6092.

\bibitem[{Ma et~al.(2021)Ma, Ethayarajh, Thrush, Jain, Wu, Jia, Potts,
  Williams, and Kiela}]{ma2021dynaboard}
Zhiyi Ma, Kawin Ethayarajh, Tristan Thrush, Somya Jain, Ledell Wu, Robin Jia,
  Christopher Potts, Adina Williams, and Douwe Kiela. 2021.
\newblock \href
  {https://proceedings.neurips.cc/paper/2021/hash/55b1927fdafef39c48e5b73b5d61ea60-Abstract.html}
  {Dynaboard: An evaluation-as-a-service platform for holistic next-generation
  benchmarking}.
\newblock \emph{Advances in Neural Information Processing Systems},
  34:10351--10367.

\bibitem[{Mart{\'\i}n-Morat{\'o} et~al.(2021)Mart{\'\i}n-Morat{\'o}, Harju, and
  Mesaros}]{martin2021crowdsourcing}
Irene Mart{\'\i}n-Morat{\'o}, Manu Harju, and Annamaria Mesaros. 2021.
\newblock \href {https://ieeexplore.ieee.org/abstract/document/9632761}
  {Crowdsourcing strong labels for sound event detection}.
\newblock In \emph{2021 IEEE Workshop on Applications of Signal Processing to
  Audio and Acoustics (WASPAA)}, pages 246--250. IEEE.

\bibitem[{Masciari et~al.(2020)Masciari, Moscato, Picariello, and
  Sperl{\'\i}}]{masciari2020detecting}
Elio Masciari, Vincenzo Moscato, Antonio Picariello, and Giancarlo Sperl{\'\i}.
  2020.
\newblock \href {https://doi.org/10.1145/3410566.3410599} {Detecting fake news
  by image analysis}.
\newblock In \emph{Proceedings of the 24th symposium on international database
  engineering \& Applications}, pages 1--5.

\bibitem[{McCrae et~al.(2022)McCrae, Wang, and Zakhor}]{mccrae2022multi}
Scott McCrae, Kehan Wang, and Avideh Zakhor. 2022.
\newblock \href
  {https://link.springer.com/chapter/10.1007/978-3-030-98355-0_28} {Multi-modal
  semantic inconsistency detection in social media news posts}.
\newblock In \emph{MultiMedia Modeling: 28th International Conference, MMM
  2022, Phu Quoc, Vietnam, June 6--10, 2022, Proceedings, Part II}, pages
  331--343. Springer.

\bibitem[{Paun et~al.(2018)Paun, Carpenter, Chamberlain, Hovy, Kruschwitz, and
  Poesio}]{paun2018comparing}
Silviu Paun, Bob Carpenter, Jon Chamberlain, Dirk Hovy, Udo Kruschwitz, and
  Massimo Poesio. 2018.
\newblock \href {https://doi.org/10.1162/tacl_a_00040} {Comparing bayesian
  models of annotation}.
\newblock \emph{Transactions of the Association for Computational Linguistics},
  6:571--585.

\bibitem[{Qi et~al.(2023)Qi, Bu, Cao, Ji, Shui, Xiao, Wang, and Chua}]{fakesv}
Peng Qi, Yuyan Bu, Juan Cao, Wei Ji, Ruihao Shui, Junbin Xiao, Danding Wang,
  and Tat-Seng Chua. 2023.
\newblock \href {https://arxiv.org/pdf/2211.10973v2.pdf} {Fakesv: A multimodal
  benchmark with rich social context for fake news detection on short video
  platforms}.
\newblock In \emph{Proceedings of the AAAI Conference on Artificial
  Intelligence}. AAAI.

\bibitem[{Rony et~al.(2017)Rony, Hassan, and Yousuf}]{rony2017diving}
Md~Main~Uddin Rony, Naeemul Hassan, and Mohammad Yousuf. 2017.
\newblock \href {https://dl.acm.org/doi/abs/10.1145/3110025.3110054} {Diving
  deep into clickbaits: Who use them to what extents in which topics with what
  effects?}
\newblock In \emph{Proceedings of the 2017 IEEE/ACM international conference on
  advances in social networks analysis and mining 2017}, pages 232--239.

\bibitem[{Samory et~al.(2020)Samory, Abnousi, and
  Mitra}]{samory2020characterizing}
Mattia Samory, Vartan~Kesiz Abnousi, and Tanushree Mitra. 2020.
\newblock \href {https://doi.org/10.1609/icwsm.v14i1.7327} {Characterizing the
  social media news sphere through user co-sharing practices}.
\newblock In \emph{Proceedings of the International AAAI Conference on Web and
  Social Media}, volume~14, pages 602--613.

\bibitem[{Sandri et~al.(2023)Sandri, Leonardelli, Tonelli, and
  Jezek}]{sandri-etal-2023-dont}
Marta Sandri, Elisa Leonardelli, Sara Tonelli, and Elisabetta Jezek. 2023.
\newblock \href {https://aclanthology.org/2023.eacl-main.178} {Why don{'}t you
  do it right? analysing annotators{'} disagreement in subjective tasks}.
\newblock In \emph{Proceedings of the 17th Conference of the European Chapter
  of the Association for Computational Linguistics}, pages 2428--2441,
  Dubrovnik, Croatia. Association for Computational Linguistics.

\bibitem[{Shang et~al.(2019)Shang, Zhang, Wang, Lai, and
  Wang}]{shang2019towards}
Lanyu Shang, Daniel~Yue Zhang, Michael Wang, Shuyue Lai, and Dong Wang. 2019.
\newblock \href {https://doi.org/10.1016/j.knosys.2019.07.022} {Towards
  reliable online clickbait video detection: A content-agnostic approach}.
\newblock \emph{Knowledge-Based Systems}, 182:104851.

\bibitem[{Snow et~al.(2008)Snow, O{'}Connor, Jurafsky, and
  Ng}]{snow-etal-2008-cheap}
Rion Snow, Brendan O{'}Connor, Daniel Jurafsky, and Andrew Ng. 2008.
\newblock \href {https://aclanthology.org/D08-1027} {Cheap and fast {--} but is
  it good? evaluating non-expert annotations for natural language tasks}.
\newblock In \emph{Proceedings of the 2008 Conference on Empirical Methods in
  Natural Language Processing}, pages 254--263, Honolulu, Hawaii. Association
  for Computational Linguistics.

\bibitem[{Song et~al.(2016)Song, Redi, Vallmitjana, and Jaimes}]{song2016click}
Yale Song, Miriam Redi, Jordi Vallmitjana, and Alejandro Jaimes. 2016.
\newblock \href {https://doi.org/10.1145/2983323.2983349} {To click or not to
  click: Automatic selection of beautiful thumbnails from videos}.
\newblock In \emph{Proceedings of the 25th ACM international on conference on
  information and knowledge management}, pages 659--668.

\bibitem[{Starbird et~al.(2019)Starbird, Arif, and
  Wilson}]{starbird2019disinformation}
Kate Starbird, Ahmer Arif, and Tom Wilson. 2019.
\newblock \href {https://doi.org/10.1145/3359229} {Disinformation as
  collaborative work: Surfacing the participatory nature of strategic
  information operations}.
\newblock \emph{Proceedings of the ACM on Human-Computer Interaction},
  3(CSCW):1--26.

\bibitem[{Thomason et~al.(2019)Thomason, Gordon, and Bisk}]{thomason2019single}
Jesse Thomason, Daniel Gordon, and Yonatan Bisk. 2019.
\newblock \href {https://doi.org/10.18653/v1/N19-1197} {Shifting the baseline:
  Single modality performance on visual navigation {\&} {QA}}.
\newblock In \emph{Proceedings of the 2019 Conference of the North {A}merican
  Chapter of the Association for Computational Linguistics: Human Language
  Technologies, Volume 1 (Long and Short Papers)}, pages 1977--1983,
  Minneapolis, Minnesota. Association for Computational Linguistics.

\bibitem[{Toledo et~al.(2019)Toledo, Gretz, Cohen-Karlik, Friedman, Venezian,
  Lahav, Jacovi, Aharonov, and Slonim}]{toledo2019automatic}
Assaf Toledo, Shai Gretz, Edo Cohen-Karlik, Roni Friedman, Elad Venezian, Dan
  Lahav, Michal Jacovi, Ranit Aharonov, and Noam Slonim. 2019.
\newblock \href {https://aclanthology.org/D19-1564} {Automatic argument quality
  assessment-new datasets and methods}.
\newblock In \emph{Proceedings of the 2019 Conference on Empirical Methods in
  Natural Language Processing and the 9th International Joint Conference on
  Natural Language Processing (EMNLP-IJCNLP)}, pages 5625--5635.

\bibitem[{Vosoughi et~al.(2018)Vosoughi, Roy, and Aral}]{vosoughi2018spread}
Soroush Vosoughi, Deb Roy, and Sinan Aral. 2018.
\newblock \href {https://www.science.org/doi/full/10.1126/science.aap9559} {The
  spread of true and false news online}.
\newblock \emph{science}, 359(6380):1146--1151.

\bibitem[{Wakefield(2016)}]{wakefield}
Jane Wakefield. 2016.
\newblock \href {https://www.bbc.com/news/uk-36528256} {Social media 'outstrips
  tv' as news source for young people}.
\newblock \emph{BBC News}.

\bibitem[{Walker and Matsa(2021)}]{walker}
Mason Walker and Katerina~Eva Matsa. 2021.
\newblock \href
  {https://www.pewresearch.org/journalism/2021/09/20/news-consumption-across-social-media-in-2021/}
  {News consumption across social media in 2021}.
\newblock \emph{Pew Research Center}.

\bibitem[{Wallace et~al.(2019)Wallace, Rodriguez, Feng, Yamada, and
  Boyd-Graber}]{wallace2019trick}
Eric Wallace, Pedro Rodriguez, Shi Feng, Ikuya Yamada, and Jordan Boyd-Graber.
  2019.
\newblock \href {https://aclanthology.org/Q19-1029} {Trick me if you can:
  Human-in-the-loop generation of adversarial examples for question answering}.
\newblock \emph{Transactions of the Association for Computational Linguistics},
  7:387--401.

\bibitem[{Wang et~al.(2021)Wang, Pang, and Pavlou}]{wang2021seeing}
Shuting~Ada Wang, Min-Seok Pang, and Paul~A Pavlou. 2021.
\newblock \href {https://papers.ssrn.com/sol3/papers.cfm?abstract_id=3909942}
  {Seeing is believing? how including a video in fake news influences users'
  reporting the fake news to social media platforms}.
\newblock \emph{How Including a Video in Fake News Influences Users' Reporting
  the Fake News to Social Media Platforms (August 23, 2021)}.

\bibitem[{Wang et~al.(2019)Wang, McKee, Torbica, and
  Stuckler}]{wang2019systematic}
Yuxi Wang, Martin McKee, Aleksandra Torbica, and David Stuckler. 2019.
\newblock \href {https://doi.org/10.1016/j.socscimed.2019.112552} {Systematic
  literature review on the spread of health-related misinformation on social
  media}.
\newblock \emph{Social science \& medicine}, 240:112552.

\bibitem[{Wei and Wan(2017)}]{wei2017learning}
Wei Wei and Xiaojun Wan. 2017.
\newblock \href {https://dl.acm.org/doi/abs/10.5555/3171837.3171869} {Learning
  to identify ambiguous and misleading news headlines}.
\newblock In \emph{Proceedings of the 26th International Joint Conference on
  Artificial Intelligence}, pages 4172--4178.

\bibitem[{Xu et~al.(2021{\natexlab{a}})Xu, Ghosh, Huang, Arora, Aminzadeh,
  Feichtenhofer, Metze, and Zettlemoyer}]{xu2021vlm}
Hu~Xu, Gargi Ghosh, Po-Yao Huang, Prahal Arora, Masoumeh Aminzadeh, Christoph
  Feichtenhofer, Florian Metze, and Luke Zettlemoyer. 2021{\natexlab{a}}.
\newblock \href {https://aclanthology.org/2021.findings-acl.370} {Vlm:
  Task-agnostic video-language model pre-training for video understanding}.
\newblock In \emph{Findings of the Association for Computational Linguistics:
  ACL-IJCNLP 2021}, pages 4227--4239.

\bibitem[{Xu et~al.(2021{\natexlab{b}})Xu, Ghosh, Huang, Okhonko, Aghajanyan,
  Metze, Zettlemoyer, and Feichtenhofer}]{xu2021videoclip}
Hu~Xu, Gargi Ghosh, Po-Yao Huang, Dmytro Okhonko, Armen Aghajanyan, Florian
  Metze, Luke Zettlemoyer, and Christoph Feichtenhofer. 2021{\natexlab{b}}.
\newblock \href {https://aclanthology.org/2021.emnlp-main.544} {Videoclip:
  Contrastive pre-training for zero-shot video-text understanding}.
\newblock In \emph{Proceedings of the 2021 Conference on Empirical Methods in
  Natural Language Processing}, pages 6787--6800.

\bibitem[{Yang et~al.(2015)Yang, Yih, and Meek}]{yang2015wikiqa}
Yi~Yang, Wen-tau Yih, and Christopher Meek. 2015.
\newblock \href {https://aclanthology.org/D15-1237} {Wikiqa: A challenge
  dataset for open-domain question answering}.
\newblock In \emph{Proceedings of the 2015 conference on empirical methods in
  natural language processing}, pages 2013--2018.

\bibitem[{You et~al.(2023)You, Lin, Lin, and Cao}]{you2023video}
Jinpeng You, Yanghao Lin, Dazhen Lin, and Donglin Cao. 2023.
\newblock \href
  {https://link.springer.com/chapter/10.1007/978-3-030-98355-0_28} {Video rumor
  classification based on multi-modal theme and keyframe fusion}.
\newblock In \emph{Computer Supported Cooperative Work and Social Computing:
  17th CCF Conference, ChineseCSCW 2022, Taiyuan, China, November 25--27, 2022,
  Revised Selected Papers, Part I}, pages 58--72. Springer.

\bibitem[{Zaidan et~al.(2007)Zaidan, Eisner, and
  Piatko}]{zaidan-etal-2007-using}
Omar Zaidan, Jason Eisner, and Christine Piatko. 2007.
\newblock \href {https://aclanthology.org/N07-1033} {Using {``}annotator
  rationales{''} to improve machine learning for text categorization}.
\newblock In \emph{Human Language Technologies 2007: The Conference of the
  North {A}merican Chapter of the Association for Computational Linguistics;
  Proceedings of the Main Conference}, pages 260--267, Rochester, New York.
  Association for Computational Linguistics.

\bibitem[{Zannettou et~al.(2018)Zannettou, Chatzis, Papadamou, and
  Sirivianos}]{zannettou2018good}
Savvas Zannettou, Sotirios Chatzis, Kostantinos Papadamou, and Michael
  Sirivianos. 2018.
\newblock \href {https://ieeexplore.ieee.org/document/8424634} {The good, the
  bad and the bait: Detecting and characterizing clickbait on youtube}.
\newblock In \emph{2018 IEEE Security and Privacy Workshops (SPW)}, pages
  63--69. IEEE.

\bibitem[{Zimdars(2016)}]{Zimdars}
Melissa Zimdars. 2016.
\newblock \href
  {https://www.washingtonpost.com/posteverything/wp/2016/11/18/my-fake-news-list-went-viral-but-made-up-stories-are-only-part-of-the-problem/}
  {My ‘fake news list’ went viral. but made-up stories are only part of the
  problem.}
\newblock \emph{The Washington Post}.

\end{thebibliography}
\bibliographystyle{style/acl_natbib}

\appendix
\clearpage
\appendix
\section{Selection of Venues} \label{venuesource} 
We selected videos introduced by \citet{rony2017diving} where the videos were created by mainstream media consisting of 25 most circulated print media and 43 most-watched broadcast media , and unreliable media cross-checked by two sources, 
informationbeautiful\footnote{\href{https://docs.google.com/spreadsheets/d/1xDDmbr54qzzG8wUrRdxQl_C1dixJSIYqQUaXVZBqsJs/edit?usp=sharing}{Unreliable Fake News Sites}} 
and \citet{Zimdars} in the US. These were selected to include a broad range of media outlets that may include misinformation. 

\section{Annotation Task}

\begin{figure}[!t]
    \centering
    \includegraphics[width=\linewidth]{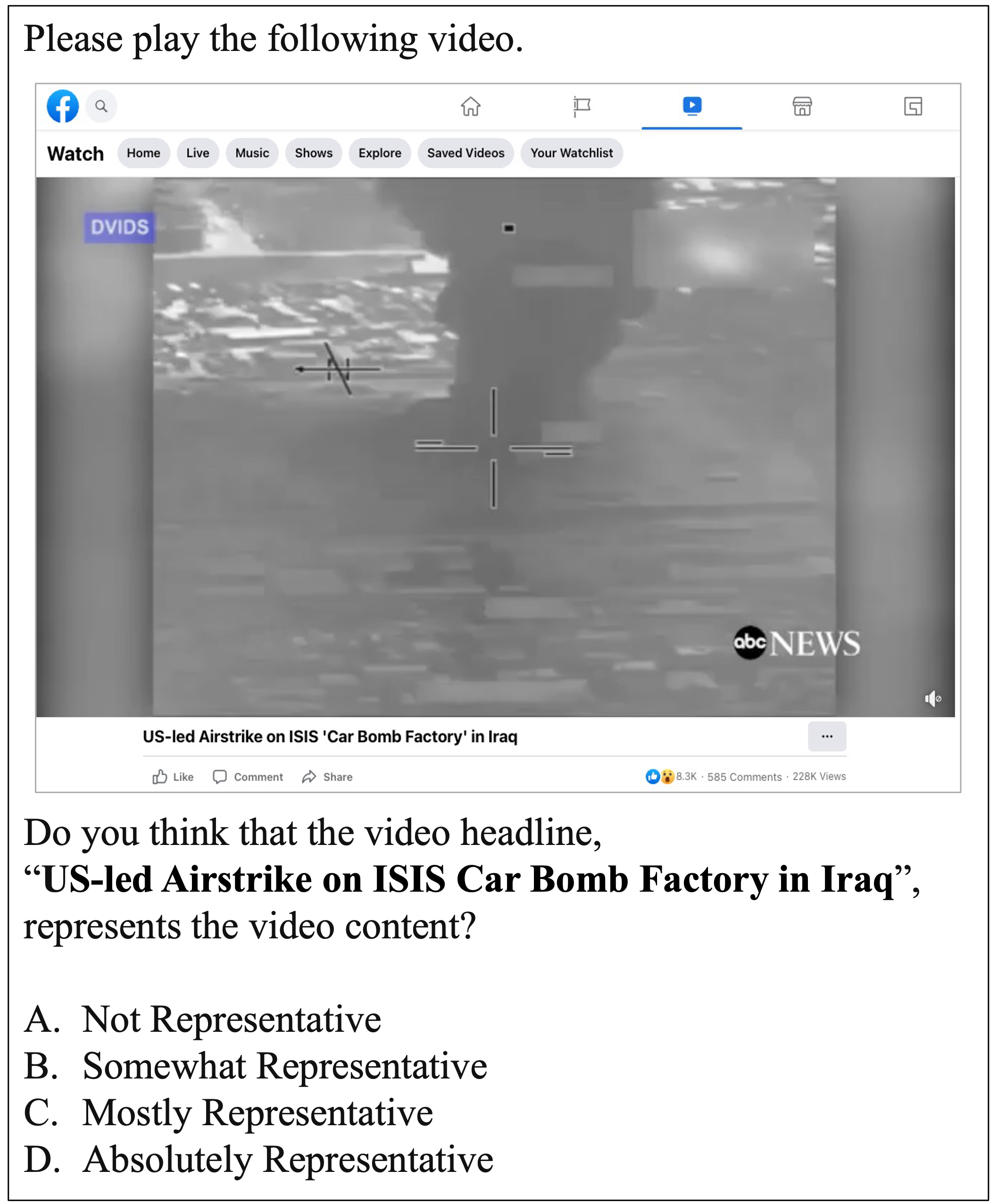}
    \caption{Example of Pilot Study. The word "represents" was too ambiguous for the audience, causing the annotators to interpret the task differently; thus it was difficult for them to consider the misleadingness of a headline.}
    \label{fig:previous}
\end{figure}

\paragraph{Example of Pilot Study} \label{pilotstudy}
As demonstrated in Figure \ref{fig:previous}, our pilot study revealed that asking one question whether the video headline represented the video caused much confusion around the word \textit{represents}, making it too ambiguous for the workers to answer the question properly. After a few interactions with workers, we found that this was due to the inherent subjectivity of the \textit{Misleading Video Headline Detection} Task. 

\section{Questions for Headline Property}\label{Appendix:headline_property_questions}
We found out from a preliminary survey that merely asking a question, \emph{how well do you think the video headline represents the video content} causes confusion among workers due to the question's inherent subjectivity. We assume that for different types of headlines, people follow different cognitive processes when assessing the headline's misleadingness. Thus, we first assess the properties of the headline and ask the following questions. Examples are in Table \ref{table:statement_example} and Table \ref{table:question_example}.

\begin{table*}[h]
    \centering
    \small
    \setlength{\tabcolsep}{5pt} 
    \renewcommand{\arraystretch}{1.3}
    \begin{tabular}{ccc}
     \Xhline{1.5pt}
        \rowcolor{Gray}
        \textbf{Factual} & \textbf{Opinionated} & \textbf{Neither} \\ 
        \rowcolor{Gray}
        \textbf{Statement} &   \textbf{Statement} & \textbf{Statement} \\
         \Xhline{1.5pt}
        \multirow{2}{*}{Biden was not elected in 2020}  &    Best ways to make oatmeal   &  \multirow{2}{*}{Great Depression} \\ 
          &    (The word ‘best’ is open to interpretation)  &  \\
        \multirow{2}{*}{Trump has 10 children}   & The power of healthy food  &      \multirow{2}{*}{Make your own coconut milk} \\
           & (The word ‘healthy’ is open to interpretation)  &      \\
       \multirow{2}{*}{She provided tips for making oatmeal}   & Vulgar language from Trump &   \multirow{2}{*}{Tips for making oatmeal} \\
       & (The word ‘vulgar’ is open to interpretation)  &  \\
       \multirow{3}{*}{Trump to Biden: 'You're the Puppet'}   & 5 minutes of truth &   \multirow{3}{*}{Trump’s wife} \\
       & (The word ‘truth’ may imply different   &  \\
       & things depending on your experience) & \\
        \Xhline{1.5pt}
    \end{tabular}
    \caption{Examples for Selecting Statement Headline Categories}
    \label{table:statement_example}
\end{table*}

\begin{table*}[h]
    \centering
    \small
    \setlength{\tabcolsep}{5pt} 
    \renewcommand{\arraystretch}{1.3}
    \begin{tabular}{cc}
     \Xhline{1.5pt}
        \rowcolor{Gray}
        \textbf{Factual Question} & \textbf{Opinionated Question} \\ 
         \Xhline{1.5pt}
        \multirow{2}{*}{Did Trump win the election?}  &    VP debate: Do you want a ``you're hired" president?    \\ 
          &    (The question is asking for your personal preference)  \\[1.2ex]
        \multirow{2}{*}{ When were the first automobiles invented?}   & What started the French revolution?  \\
           & (The question is asking something that is open to different interpretations) \\[1.2ex]
       \multirow{2}{*}{Do you check the temperature every day?}   & What if I made you eat worms? \\
       & (The question is asking for your personal preference)  \\ [1.2ex] 
    \Xhline{1.5pt}
    \end{tabular}
    \caption{Annotators are given five headline properties to choose what kind of sentence headline is.}
    \label{table:question_example}
\end{table*}

\begin{table*}[!t]
    \centering
    \footnotesize
    \setlength{\tabcolsep}{5pt} 
    \renewcommand{\arraystretch}{1.3}
     \begin{tabular}{ccc}
     \Xhline{1.5pt}
        \rowcolor{Gray}
        \textbf{Original Headline} & \textbf{Synthesized Headlines}  & \textbf{Groundings} \\ 
        \Xhline{1.5pt}
        This \colorbox{red!15}{woman} takes some of the most  &     This \colorbox{blue!15}{man} takes some of the most  & False (because it is a ``woman" not \\
        dangerous selfies in the world  &     dangerous selfies in the world  &  a man who is taking selfies in the video) \\ [1.2ex]
        Baby \colorbox{red!15}{Girl} Gets Adorably Upset   & Baby \colorbox{blue!15}{Boy} Gets Adorably  &      False (because it is a ``girl" not \\
        When Parents Kiss In Front Of Her   & When Parents Kiss In Front Of Him  &   a boy who cries in the video)\\[1.2ex]
        \multirow{2}{*}{Trump to \colorbox{red!15}{Clinton}: 'You're the Puppet'} & \multirow{2}{*}{Trump to \colorbox{blue!15}{Biden}: 'You're the Puppet'}  & False (because It is ``Clinton" not \\ 
        & &  Biden that counters Trump in the video) \\[1.2ex]
        \multirow{2}{*}{\colorbox{red!15}{Toyota} created a mini robot companion} & \multirow{2}{*}{\colorbox{blue!15}{Honda} created a mini robot companion}  & False (because It is ``Toyota" not  \\ & &  Honda mentioned in the video) \\ 
        \Xhline{1.5pt}
    \end{tabular}
    \caption{\textbf{Examples of Synthesized Headlines for Accuracy-check Questions}}
    \label{Table:accuracychecktable}
\end{table*}

\paragraph{Opinionated Statement} If the worker chooses that a given headline is a \emph{opinionated statement}, the consecutive question would be \emph{Do you have prior knowledge about the statement in the headline to make a judgment on the statement?} to assess their original opinion stated in the headline. After watching the video, the workers are asked \emph{\bf{Assuming that the information provided by the video is correct}, how would you rate the following statement? \bf{The video justifies the opinion in the headline.}} This question specifically asks to find the congruence between the video's message and the opinion stated in the headline. If the worker finds the video content appropriate enough to match the headline, they are expected to select \emph{Agree}. Then we conclude that the final label of the video headline is \emph{\representative}.

\paragraph{Neither Statement} If the worker chooses that a given headline is a \emph{neither statement}, the consecutive question would be \emph{Write down what you expect to see in a video} to assess their background knowledge about the headline and what they expect to see in the video. After watching the video, the workers are asked \emph{\bf{Assuming that the information provided by the video is correct}, how would you rate the following statement? \bf{The video talks about the video.}} This question specifically asks to find the congruence between the video's message and the information in the headline. If the worker finds the video content appropriate enough to match the headline, they are expected to select \emph{Agree}. Then we conclude that the final label of the video headline is \emph{\representative}.

\paragraph{Factual/Opinionated Question} If the worker chooses that a given headline is in the form of \emph{question}, we ask the same questions for both factual and opinionated questions. Before watching the video, the consecutive question would be \emph{Write down what you expect to see in a video} to assess their background knowledge about the headline and what they expect to see in the video. After watching the video, the workers are asked \emph{\bf{Assuming that the information provided by the video is correct}, how would you rate the following statement? \bf{The information provided by the video helps you answer the question in the headline.}} This question specifically asks to find an answer to the question in the headline, assuming that video content is expected to contain the information that the headline is inquiring about. If the worker decides that the video content cannot answer or has insufficient information, they are expected to select \emph{Disagree}. Then we conclude that the final label of the video headline is \emph{misleading}.

\section{Quality Control and Assessment}
\paragraph{Pre-qualification Test}
We restrict this task to the workers in the United States given that they have a higher possibility of being fluent in the verbal and literal understanding of English. Before the workers participate in the HIT, we prepare a preliminary qualification test that the workers must pass to start the HIT. All the participants must take this pre-qualification test, given multi-choice questions such as “How \emph{representative} is the video?” and “How would you rewrite the headline.” When they receive a score of 100, they are qualified to participate in the following HITs. This process is included to ensure that the participants have the capacity to integratively comprehend the video content and video headline, and then draw out an accurate video label. 

\paragraph{Synthesized Headlines in Accuracy Check Questions} \label{accuracycheck}
Table \ref{Table:accuracychecktable} shows examples of synthesized headlines in accuracy check questions. Accuracy check questions that are synthetically created to be always misleading (obviously false). For each annotator, we calculate the ratio between the number of correct answers and the number of accuracy check questions to select competent annotators.

\paragraph{MACE}\label{Appendix:MACE}
We compute MACE, a Bayesian approach-based metric that takes into account the credibility of the annotator and their spamming preference \citep{hovy2013learning}. 
\begin{align}
    \nonumber & \text{for} ~ i=1,\cdots,N:  \\
    \nonumber & \quad\quad T_{i} \sim \text{Uniform}   \\
    \nonumber & \quad\quad \text{for} ~ j=1,\cdots,M : \\
    \nonumber & \quad\quad\quad\quad S_{ij} \sim \text{Bernoulli}(1-\theta_{j}) \\
    \nonumber & \quad\quad\quad\quad \text{if} ~ S_{ij} = 0 :\\
    \nonumber & \quad\quad\quad\quad\quad\quad A_{ij} = T_{i} \\
    \nonumber & \quad\quad\quad\quad \text{else} : \\
    \nonumber & \quad\quad\quad\quad\quad\quad A_{ij} \sim \text{Multinomial}(\xi_{j}),
\end{align}
where $N$ denotes the number of headlines, $T$ denotes the number of the true labels, and $M$ denotes the number of workers. $S_{ij}$ denotes the spam indicator of worker $j$ for annotating headline $i$, while $A_{ij}$ denotes the annotation of worker $j$ for headline $i$. $\theta$ and $\xi$ each denotes the parameter of worker $j$’s trustworthiness and spam pattern. We add Beta and Dirichlet priors on $\theta$ and $\xi$ respectively. The assumption in the generative process is that an annotator always produces the correct label when he does not show a spam pattern which helps in excluding the labels that are not correlated with the correct label. Here, our parameter of interest is $\theta$ which stands for the trustworthiness of each worker. We apply \citet{paun2018comparing}’s implementation to obtain posterior distributions (samples) of $\theta$ and calculate point estimates.

\section{Other Feature Distribution} \label{metadata}
The venue kind \emph{Website} show higher percentage (29\%) of creating misleading headlines (Table \ref{tab:venue_kind_dist}). On the other hand, because the proportions of misleading headlines are fairly uniform in the
1) proportions of news topics, 2) headline properties, and 3) venue credibility, it suggests that the three features are less prone to be an indicator for misleading headlines (The proportions of each label in the three features are reported in Table \ref{tab:topic_dist}, \ref{tab:property_dist} and \ref{tab:venue_cred_dist}). 

\begin{table}[!t]
    \centering
    \begin{tabular}{cccc}
    \Xhline{1.5pt}
    \rowcolor{Gray}
    & \multicolumn{2}{c}{Annotated Labels}\\
    \rowcolor{Gray}
          \multirow{-2}{*}{\textbf{Venue Kind}}     
          &{Representative} & {Misleading}
         \\ 
         \Xhline{1.5pt}
         Broadcast    & 0.85 & 0.15 \\
         Cable          & 0.85 & 0.15 \\
         Newspaper      & 0.87 & 0.13 \\
         Website       & 0.71 & 0.29 \\
    \Xhline{1.5pt}
    \end{tabular}
    \caption{\emph{Website} shows more proportion of creating misleading headlines than other categories in the venue kind feature, which suggests that venue kind feature may be an indicator of representativeness of a headline.}
    \label{tab:venue_kind_dist}
\end{table}

\begin{table}[!t]
    \centering
    \begin{tabular}{cccc}
    \Xhline{1.5pt}
    \rowcolor{Gray}
    & \multicolumn{2}{c}{Annotated Labels}\\
    \rowcolor{Gray}
          \multirow{-2}{*}{\textbf{Headline Topics}}     
          &{Representative} & {Misleading}
         \\ 
         \Xhline{1.5pt}
         {Entertainment}    & 0.86 & 0.14 \\
         Food                & 0.86 & 0.14 \\
         Others                 & 0.81 & 0.19 \\
         Politics                  & 0.85 & 0.15 \\
    \Xhline{1.5pt}
    \end{tabular}
    \caption{There was no significant difference in the proportions of topics, which suggests that topic feature is not strong indicator for misleadingness.}
    \label{tab:topic_dist}
\end{table}

\begin{table}[!t]
    \centering
    \scalebox{0.9}{
    \begin{tabular}{cccc}
    \Xhline{1.5pt}
    \rowcolor{Gray}
    & \multicolumn{2}{c}{Annotated Labels}\\
    \rowcolor{Gray}
          \multirow{-2}{*}{\textbf{Headline Properties}}     
          &{Representative} & {Misleading}
         \\ 
         \Xhline{1.5pt}
         {Factual Statement}    & 0.86 & 0.14 \\
         Opinionated Statement  & 0.84 & 0.16 \\
         Neither Statement      & 0.83 & 0.17 \\
         Factual Question       & 0.81 & 0.19 \\
         Opinionated Question   & 0.72 & 0.28 \\
    \Xhline{1.5pt}
    \end{tabular}}
    \caption{There was no significant difference in the proportions of properties, which suggests that property feature is not strong indicator for misleadingness.}
    \label{tab:property_dist}
\end{table}

\begin{table}[!t]
    \centering
    \scalebox{0.9}{
    \begin{tabular}{cccc}
    \Xhline{1.5pt}
    \rowcolor{Gray}
    & \multicolumn{2}{c}{Annotated Labels}\\
    \rowcolor{Gray}
          \multirow{-2}{*}{\textbf{Venue Credibility}}     
          &{Representative} & {Misleading}
         \\ 
         \Xhline{1.5pt}
         {High}    & 0.86 & 0.14 \\
         Mostly Factual  & 0.84 & 0.16 \\
         Mixed      & 0.85 & 0.15 \\
         Low       & 0.81 & 0.19 \\
         Unknown   & 0.85 & 0.15 \\
    \Xhline{1.5pt}
    \end{tabular}}
    \caption{There was no significant difference in the proportions of properties, which suggests that the headline property feature is not strong indicator for misleadingness.}
    \label{tab:venue_cred_dist}
\end{table}

\section{What Makes for Misleadingness in Rationales?} \label{subrationales}
\begin{figure}[!t]
    \centering
    \includegraphics[width=\linewidth]{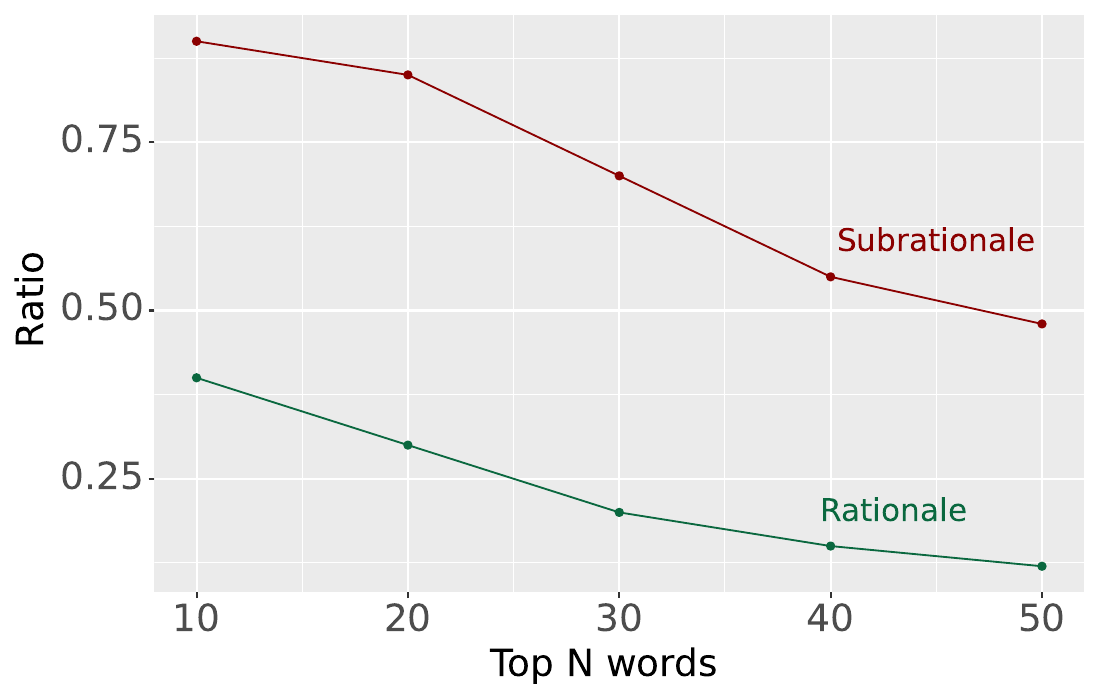}
    \caption{The top N words selected from the Random Forest Classifier to predict the correct label were mostly included in subrationales compared to rationales. As N increases, the ratio of overlapping words between the subrationale and top N important words stays higher than that of the rationale.} 
    \label{fig:rationaletype}
\end{figure}
To specifically understand how rationales help in predicting the correct \textit{misleading} class, we trained Random Forest classifier using \abr{tf-idf} features of \{Headline + Rationale + Subrationale\}. Figure \ref{fig:rationaletype} shows the ratio of overlapping words between two types of rationales and top N important words. The top 10 words selected from the Random Forest Classifier to predict the correct label were mostly included in subrationales compared to rationales (Table \ref{table:subrationale_ex}).
\section{Finetuning Details of Baseline Models} \label{finetune}

\begin{figure*}[!t]
    \centering
    \includegraphics[width=\linewidth]{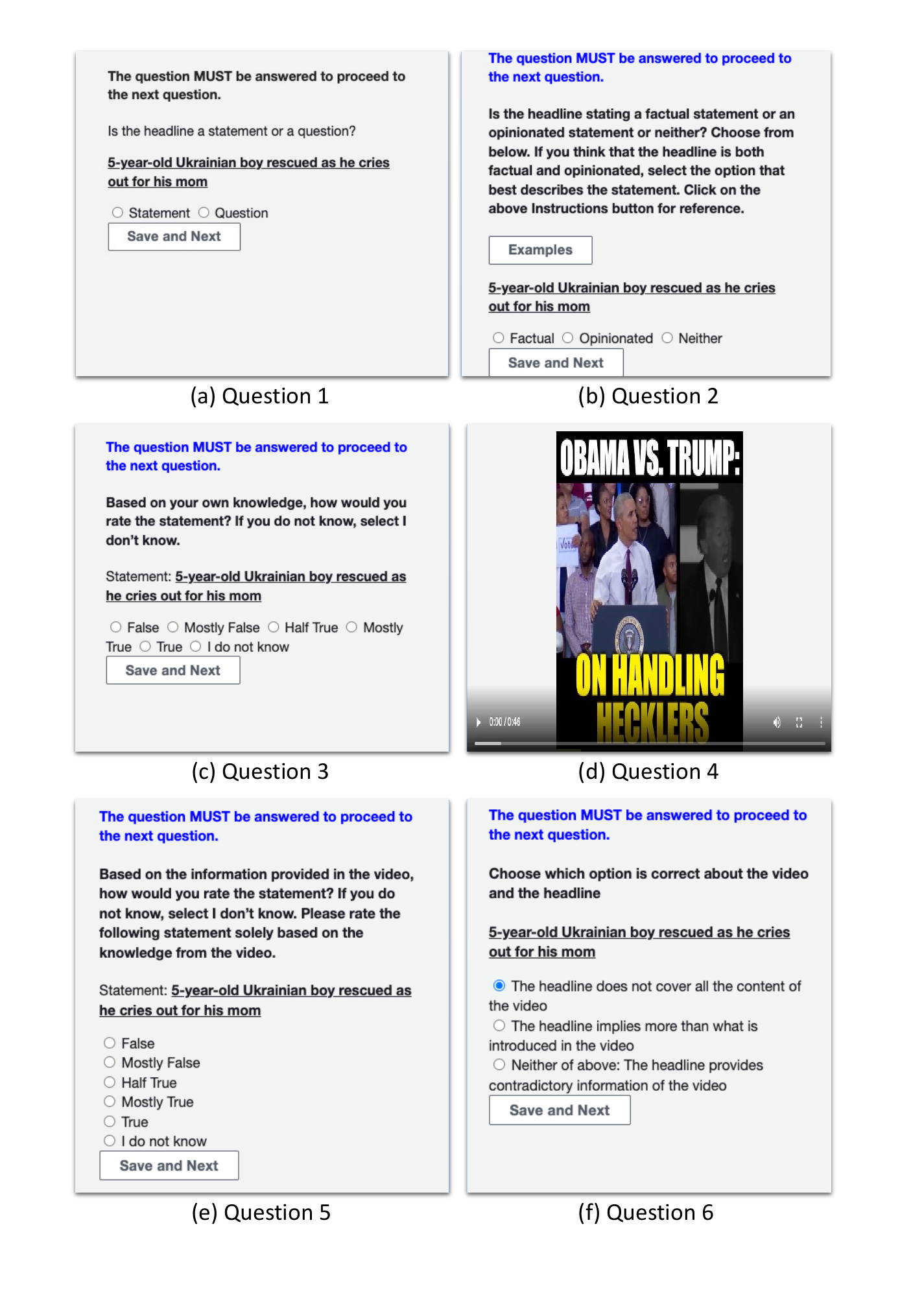}
    \caption{\textbf{Survey Example Distributed in Mturk}}
    \label{fig:survey}
\end{figure*}
We finetune both VideoCLIP and VLM on a A6000 GPU using the Adam optimizer with a learning rate 0.00002, weight decay ratio of 0.001, and batch size 8 for 10 epochs. For text encoders and video encoders, we directly use the best checkpoints from the pretrained VideoCLIP and VLM models. We concatenate encoder outputs, the pooled video and text features, and learn fully connected layer optimized with Cross Entropy loss. For partial input experiments, we assign zeros to text or video encoder inputs.

\section{Era of Fake News}
People have been using social media platforms to converse, diffuse and broadcast their ideas in recent years. However, there has been widespread concern that misinformation is increasing on social media, which causes damage to societies \citep{allcott2019trends}. Some contemporary commentators even describe the current period as ``an era of fake news'' \citep{wang2019systematic}. 

\section{Censoring Audio Transcripts} \label{transcripts}
We outsource transcript extractions from a software called Deepgram\footnote{https://deepgram.com/}. To validate its accuracy, we randomly sampled 55 videos that have transcripts and manually checked if the transcripts were accurate. These transcripts exactly matched the audio from the videos. \name{} also includes transcript information on the timeframe that indicates when each word starts and ends in the video with its confidence score. We especially paid attention to this information when censoring the transcripts.  
\begin{table*}[!t]
    \centering
    \small
    \renewcommand{\arraystretch}{1.6}
    \resizebox{\textwidth}{!}{
    \scalebox{1}{\begin{tabular}{p{30mm}p{30mm}p{40mm}p{13mm}}
    \Xhline{0.9pt}
    \rowcolor{Gray}
         \qquad\qquad\textbf{Headline} & \qquad \textbf{Rationale} & \qquad\qquad \textbf{Subrationale} & \textbf{Label} \\
    \Xhline{1pt}
        {Tennessee Beats Georgia With Hail Mary}  &  The headline does not cover all the content of the video & Some specific \colorbox{red!15}{information} from the video is not at all \colorbox{red!15}{reflected} in the headline & Misleading \\
        {President Obama Leaves For Final Overseas Trip}  &  The headline \colorbox{blue!15}{implies} more than what is introduced in the video & The headline uses an \colorbox{red!15}{excessively} definitive tone when the video is only suggesting the content & Misleading \\
        {Protesters Gather Outside Chicagos Trump Tower}  &  The headline \colorbox{blue!15}{implies} more than what is introduced in the video & Video \colorbox{red!15}shows a mob of people but does not \colorbox{red!15}{provide} \colorbox{red!15}{location} or reason for the protest. & Misleading  \\
        {Firefighters From Across US Battle Appalachian Wildfires}  & The headline \colorbox{blue!15}{implies} more than what is introduced in the video & The headline \colorbox{red!15}{exaggerates} the video content & Misleading \\
        {Tennessee Beats Georgia With Hail Mary}  & The headline does not cover all the content of the video & The headline chooses \colorbox{red!15}{specific} words that cannot be \colorbox{red!15}{supported} as \colorbox{red!15}{fact} & Misleading \\
        \Xhline{1pt}
    \end{tabular}}}
    \caption{The top 10 words selected from Random Forest Classifier to predict the correct label were mostly included in subrationales compared to rationales. The word ``implies'' was included in the rationales, while ``excessively'' and ``exaggerates'' included in subrationales pointed the model to correctly predict \textit{misleading}.}
    \label{table:subrationale_ex}
\end{table*}%

\end{document}